\documentclass[11pt]{article}
\usepackage{amsfonts}

\usepackage[preprint]{acl}

\usepackage{times}
\usepackage{latexsym}

\usepackage[T1]{fontenc}

\usepackage[utf8]{inputenc}

\usepackage{microtype}

\usepackage{inconsolata}

\usepackage{graphicx}
\usepackage{booktabs}
\usepackage{amsmath}
\usepackage{multirow}
\usepackage{colortbl} 
\definecolor{lightgray}{rgb}{0.9, 0.9, 0.9}
\usepackage{tabularx}
\usepackage{makecell}
\usepackage{longtable}
\usepackage{float}
\usepackage{hyperref}

\title{EmbeddingRWKV: State-Centric Retrieval with Reusable States}

\author{Haowen Hou\textsuperscript{*}
  \and 
  Jie Yang\textsuperscript{+}  \\
  \textsuperscript{*}Guangdong Laboratory of Artificial Intelligence and Digital Economy (SZ), Shenzhen, China \\
  \textsuperscript{+}Shenzhen Yuanshi Intelligence Co., Ltd, Shenzhen, China \\
  \texttt{houhaowen@gml.ac.cn}
  }

\begin{document}
\maketitle
\begin{abstract}
Current Retrieval-Augmented Generation (RAG) systems typically employ a traditional two-stage pipeline: an embedding model for initial retrieval followed by a reranker for refinement. 
However, this paradigm suffers from significant inefficiency due to the lack of shared information between stages, leading to substantial redundant computation.
To address this limitation, we propose \textbf{State-Centric Retrieval}, a unified retrieval paradigm that utilizes "states" as a bridge to connect embedding models and rerankers.
First, we perform state representation learning by fine-tuning an RWKV-based LLM, transforming it into \textbf{EmbeddingRWKV}, a unified model that serves as both an embedding model and a state backbone for extracting compact, reusable states.
Building upon these reusable states, we further design a state-based reranker to fully leverage precomputed information. 
During reranking, the model processes only query tokens, decoupling inference cost from document length and yielding a 5.4$\times$--44.8$\times$ speedup.
Furthermore, we observe that retaining all intermediate layer states is unnecessary; with a uniform layer selection strategy, our model maintains 98.62\% of full-model performance using only 25\% of the layers.
Extensive experiments demonstrate that State-Centric Retrieval achieves high-quality retrieval and reranking results while significantly enhancing overall system efficiency.
Code is available at \href{https://github.com/howard-hou/EmbeddingRWKV}{our GitHub repository}.
\end{abstract}

\section{Introduction}

\begin{figure}[th]
    \centering
    \includegraphics[width=\linewidth]{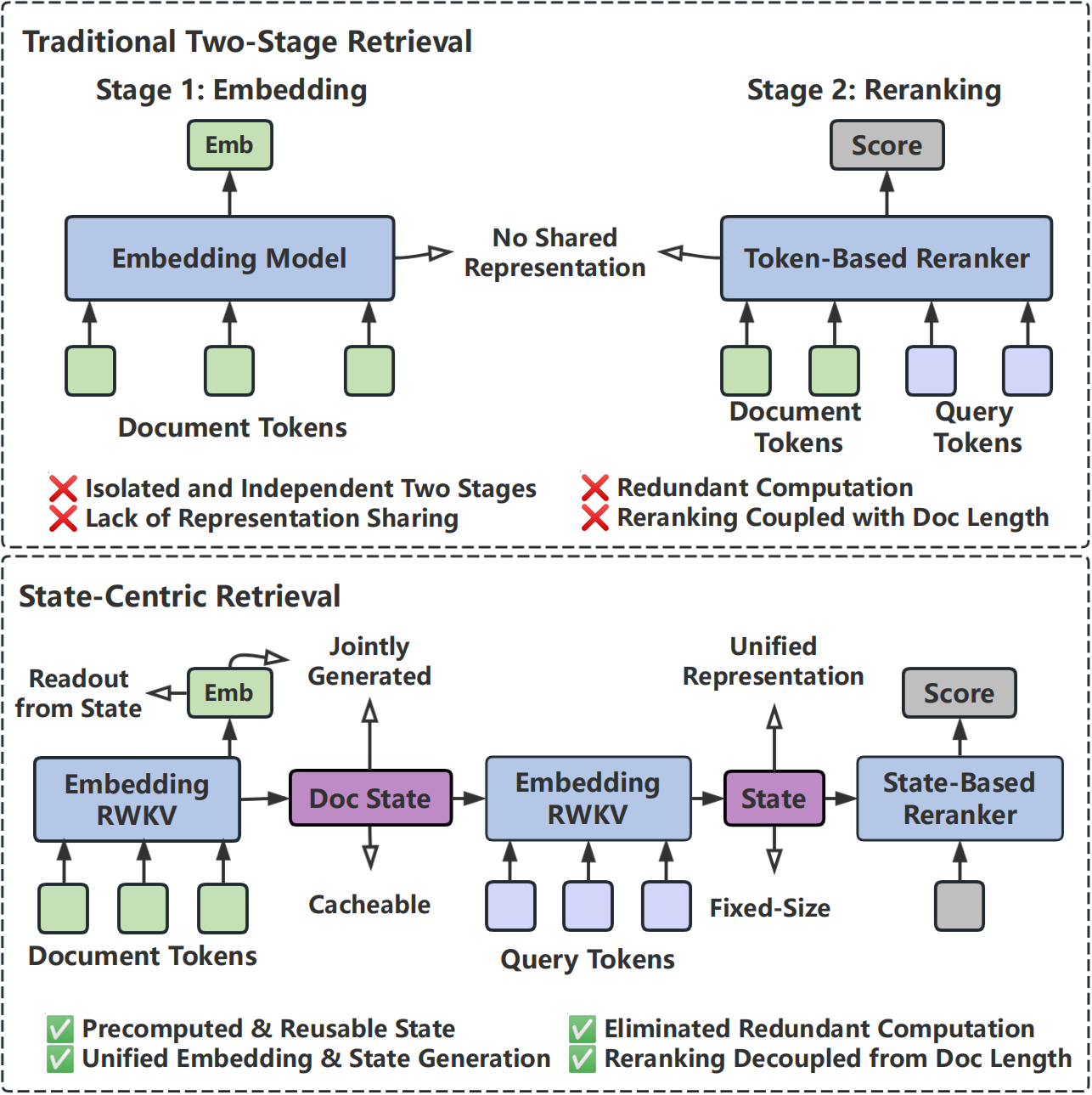}
    \caption{\textbf{Traditional vs. State-Centric Retrieval.} (Top) The traditional two-stage retrieval is fundamentally disjointed, suffering from redundant computation as the reranker re-encodes full document tokens. (Bottom) Our \textbf{State-Centric Retrieval} unifies the two stages into an efficient system via a shared, reusable state. By jointly generating embeddings and compact states, it enables offline state-based reranking that decouples inference cost from document length, \textbf{yielding 5.4$\times$--44.8$\times$ speedup}.}
    \label{fig:teaser}
\end{figure}

Retrieval-Augmented Generation (RAG) has emerged as a common paradigm in modern Large Language Models (LLMs)~\citep{brown2020language,wu2025uncovering}, enabling models to incorporate external knowledge beyond parametric memory~\citep{lewis2020retrieval}. 
Most RAG systems adopt a two-stage retrieval pipeline~\citep{glass2022re2g}: an embedding model is first used to efficiently recall candidate documents, followed by a reranker that performs refinement~\cite{li2025prorank}. 
Within this pipeline, rerankers play a critical role in improving the quality of retrieved contexts, thereby directly impacting the downstream generation performance of LLMs~\citep{zhao2025kalm}.

Despite its effectiveness, the prevailing two-stage retrieval paradigm suffers from fundamental efficiency limitations arising from both model architecture and pipeline design.
\textit{First}, most state-of-the-art embedding models and rerankers are predominantly built on Transformer architectures~\citep{qwen3embedding, jina-embedding-b-en-v1}. While powerful, Transformers incur quadratic computational complexity with respect to sequence length and exhibit rapidly growing memory consumption due to the linearly expanding key–value (KV) cache, making them increasingly inefficient in both computation and storage as context length grow.
\textit{Second}, the two-stage pipeline itself introduces additional inefficiency due to the lack of shared representations across stages as shown in Figure~\ref{fig:teaser}. 
As embedding models and rerankers operate as isolated modules, they independently encode the same documents without any information reuse, resulting in substantial redundant computation.

In this work, we propose \textbf{State-Centric Retrieval}, a novel retrieval paradigm that improves efficiency from both the architectural and pipeline perspectives.
\textit{At the model architecture level}, we adopt RWKV~\cite{peng2023rwkv,peng2024eagle,peng2025rwkv7gooseexpressivedynamic}, a linear RNN with matrix-valued hidden states that achieves linear computational complexity with respect to sequence length and constant space complexity. Compared to Transformer-based models, RWKV offers substantially improved efficiency in both computation and space.
\textit{At the pipeline level}, we introduce a unified retrieval framework (Fig.~\ref{fig:teaser}) that leverages reusable states as a shared representation across the retrieval and reranking stages, enabling information reuse and eliminating redundant computation.

We instantiate this paradigm through two synergistic contributions: \textit{learning reusable states} and \textit{efficiently utilizing these states for reranking}.

First, we observe that directly utilizing RWKV-based LLMs to extract state representations results in significant information loss. To address this, we conduct state representation learning by fine-tuning RWKV-based LLMs, transforming them into unified models that serve as both embedding models and state backbones for extracting effective state representations.
Specifically, we employ a single-stage domain-aware curriculum strategy to enhance data efficiency. This approach achieves competitive performance on MTEB benchmarks while requiring only $5\%$ of the training data typically consumed by conventional multi-stage pipelines.
This approach outperforms conventional multi-stage training pipelines on MTEB benchmarks while using only 5\% of the training data. 
Importantly, for a sequence of length $T$, the resulting states require only $32/T$ of the memory footprint of a full Transformer’s KV cache, making large-scale state caching feasible in practice.

Second, building upon cached states, we design a state-based reranker that fully exploits the precomputed information stored in states. 
During inference, the reranker processes only query tokens together with cached document states, completely decoupling inference cost from document length. 
Furthermore, we observe that retaining states from all intermediate layers is unnecessary. 
With a uniform layer selection strategy, our model preserves 98.62\% of full-model performance while using only 25\% of the layers, enabling further reductions in memory footprint and inference latency.

In summary, our contributions are as follows:
\begin{itemize}
    \item We propose \textbf{State-Centric Retrieval}, a unified framework that synergizes embedding-based retrieval and reranking through reusable states, effectively eliminating redundant computation.
    \item We introduce a domain-aware curriculum strategy for state representation learning, achieving competitive performance with only $5\%$ of the standard training data.
    \item We design a state-based reranker that significantly reduces computational overhead and memory footprint while maintaining high reranking effectiveness.
\end{itemize}

\section{Related Work}
\paragraph{Two-Stage Retrieval for Text Embeddings.}
Most retrieval-augmented generation systems adopt a two-stage retrieval paradigm, where dense embedding models are used for initial retrieval and rerankers are applied for relevance refinement~\citep{reimers2019b}. 
Text embedding models have evolved from early BERT-based encoders~\citep{devlin2018, hou2024bagformer} to large language model backbones, with recent systems such as NV-Emb~\citep{lee2025a}, E5-Mistral~\citep{wang2024a}, and GTE~\citep{zhang2025b} achieving strong performance on benchmarks like MTEB~\citep{muennighoff2023a, xiao2024}.
Despite these advances, embedding models and rerankers are typically trained and deployed as independent components. 
Each stage encodes the same documents separately, without sharing representations or intermediate information, resulting in duplicated encoding and redundant computation across the retrieval pipeline. 
This design contrasts with our state-centric retrieval framework, which enables representation reuse across retrieval stages.

\paragraph{Multi-Stage Training Recipes.}
State-of-the-art embedding models commonly rely on multi-stage training pipelines that combine large-scale pre-training with successive fine-tuning stages.
These pipelines often include supervised fine-tuning and hard-negative mining to progressively refine retrieval-specific representations.
For instance, BGE~\citep{li2025} employs RetroMAE pre-training followed by contrastive learning and instruction tuning with hard negatives. 
Jina Embeddings~\citep{sturua2024} adopts a two-stage data-centric approach that combines large-scale data filtering with triplet-based training. 
Arctic-Embed~\citep{hu2025} follows a coarse-to-fine strategy, transitioning from billion-scale pre-training to high-precision fine-tuning on curated datasets.
While effective, existing multi-stage pipelines depend on massive datasets and complex training workflows, which limit data efficiency. 
By contrast, our domain-aware curriculum strategy consolidates training into a single stage with reduced data usage, enabling a simpler and more data-efficient training process.

\paragraph{From Transformers to Linear RNNs.} 
The majority of pioneering embedding models and rerankers are built on Transformer architectures~\citep{qwen3embedding, vera2025embeddinggemma}. 
However, Transformers face inherent scalability issues: \textit{quadratic computational complexity} and a \textit{linearly growing KV cache} memory footprint~\citep{vaswani2017attention}. 
These limitations make long-context retrieval and large-scale reranking prohibitively expensive. 
To address this, our work leverages RWKV, a member of the \textit{Linear RNNs with Matrix-valued States} family. 
Unlike Transformers, RWKV achieves \textit{linear computational complexity} and \textit{constant spatial complexity}, offering a more efficient architectural backbone for dense retrieval without sacrificing the expressive power required for complex semantic matching.

\paragraph{Linear RNNs with Matrix-valued States.}
While linear RNNs like Mamba~\cite{gu2023mamba} and RWKV-4~\cite{peng2023rwkv} reduce complexity to $O(N)$, they often suffer from a \textit{state capacity gap} due to compressing context into fixed-size vectors.
To bridge this gap, recent advancements have focused on scaling hidden states into matrices. 
Since the explicit transition to matrix-valued states in RWKV-5 and RWKV-6~\cite{peng2024eagle}, the architecture has demonstrated an enhanced ability to maintain rich historical context. 
Most recently, RWKV-7~\cite{peng2025rwkv7gooseexpressivedynamic} further refined this with \textit{dynamic state evolution} mechanisms. 
By adopting this architecture, our framework enables the caching of high-capacity states that are both compact and highly representative across the retrieval pipeline.

\section{State-Centric Retrieval}
State-Centric Retrieval takes states as the primary representation. 
A precise definition of the state is therefore necessary, as it forms the basis of the entire framework and directly affects model architecture and pipeline design.

\begin{figure*}[t]
    \centering
    \includegraphics[width=\textwidth]{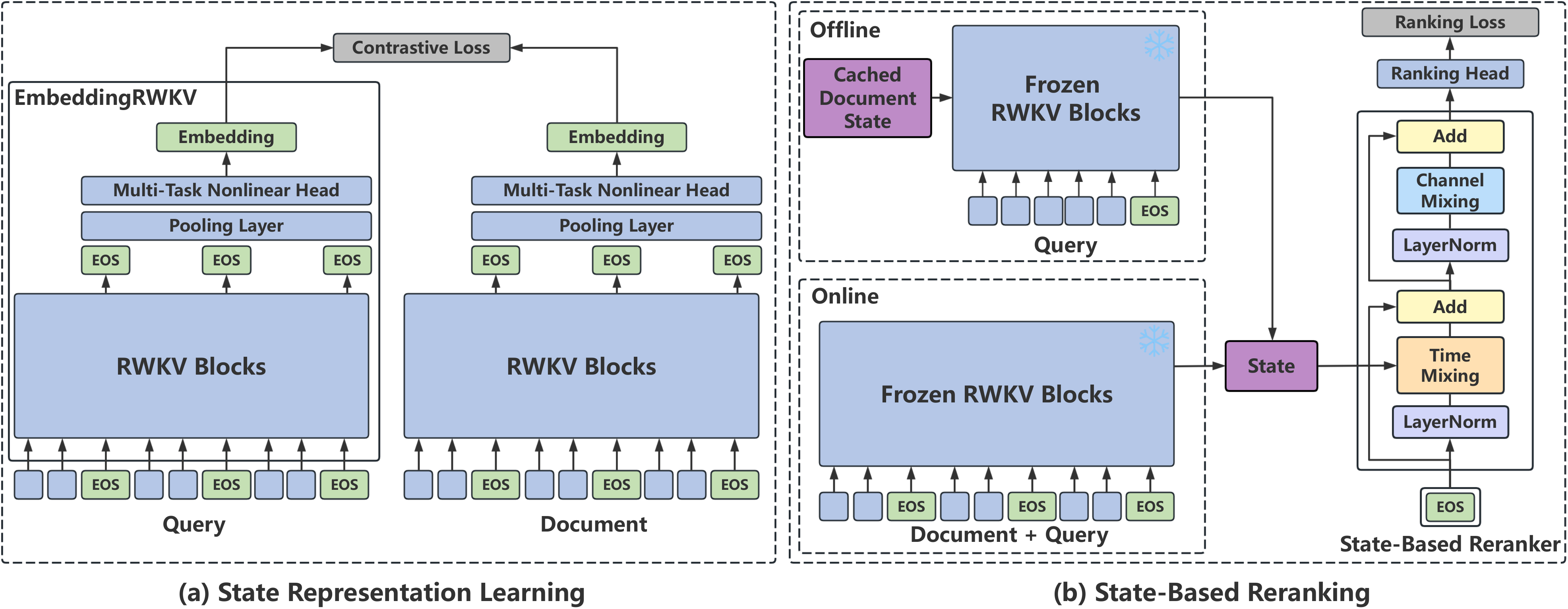}
    \caption{ 
(a) \textbf{State Representation Learning}: illustrates the process of ``Learning the State.'' 
(b) \textbf{State-Based Reranking}: demonstrates the paradigm of ``Utilizing the State'' by directly reusing the cached state for inference, thereby avoiding redundant re-computation.
}
    \label{fig:arch}
\end{figure*}

\subsection{Background: Matrix-valued States}
In recurrent models, a \textit{state} summarizes past inputs and serves as the carrier of contextual information.
Conventional RNNs~\citep{Hochreiter1997LongSM,Cho2014OnTP} represent the state as a fixed-length vector, which limits the amount of information that can be preserved over long sequences.

\textit{Matrix-valued states} generalize this formulation by representing the recurrent state as a matrix rather than a vector, allowing the model to maintain a richer and more structured memory.
Formally, given a matrix-valued state \( S_t \) and the current input \( x_t \), the recurrent update takes the form~\cite{Hou2025VisualRWKVHMEL}:
\begin{equation}
S_t = S_{t-1} W_t + v_t k_t^\top ,
\end{equation}
where $S_t\in\mathbb{R}^{d\times d}$, $W_t\in\mathbb{R}^{d\times d}$,
and $v_t, k_t \in \mathbb{R}^{d}$ are column vectors computed from $x_t$.
Under this formulation, \( S_t \) acts as a dynamic associative memory that incrementally accumulates outer products of keys and values, while selectively retaining or discarding past information through \( W_t \).

In the case of RWKV-7~\cite{peng2025rwkv7gooseexpressivedynamic}, the transition matrix $W_t$ is further refined through dynamic evolution, typically expressed by diagonalizing a vector $w_t$ and incorporating a low-rank update:
\begin{equation}
    W_t = \mathrm{diag}(w_t) - \hat{\kappa}_t \bigl(a_t \odot \hat{\kappa}_t\bigr)^\top
\end{equation}
where $\hat{\kappa}_t$ and $a_t$ are fast weights~\citep{Schlag2021LinearTA} that enhance the model's ability to selectively update its internal memory.

Importantly, the matrix-valued state can be updated incrementally in constant memory, without storing the full token history.
This property allows the model to process arbitrarily long sequences with linear-time computation and constant space complexity, while retaining strong contextual modeling capacity.

\subsection{State Representation Learning}\label{sec:method-emb}
State representation learning aims to adapt RWKV-based language models into unified backbones that can simultaneously produce high-quality embeddings and reusable states.

\paragraph{Model Architecture.}
To realize state representation learning, we introduce \emph{EmbeddingRWKV}, whose architecture is illustrated in Fig.~\ref{fig:arch}(a).
EmbeddingRWKV consists of RWKV blocks, a pooling layer, and a nonlinear head.
Multiple end-of-sequence~(EOS) tokens are inserted into the input sequence, and the corresponding output representations are extracted.
These representations are averaged by the pooling layer and projected by the nonlinear head to produce the final embedding.

\paragraph{Training and Data Recipe.}
We introduce a \emph{Domain-Aware Curriculum Strategy} for efficient state representation learning.
Data samples from the same domain share similar semantic subspaces and thus naturally form hard negatives for contrastive learning.
In contrast, randomly mixed batches across heterogeneous domains often produce easy negatives that contribute limited training signal.

To exploit this property, we organize training batches by domain in a distributed manner.
Specifically, we first partition the training dataset $\mathcal{D}$ into domain-specific subsets $\{\mathcal{D}_1, \mathcal{D}_2, \dots, \mathcal{D}_K\}$.
During distributed training on $N$ GPUs, each GPU processes a micro-batch sampled from a single domain.
Different GPUs are assigned different domains at the same training step.

As a result, samples within the same domain naturally form hard negatives within each GPU’s local batch,
This simple scheduling strategy enables effective hard-negative training without explicit mining.

\paragraph{Learning Objective.}
We train state representations using a contrastive objective, referred to as the loss $L_{\mathrm{state}}$.
we adopt a standard InfoNCE loss~\citep{oord2018representation} with in-batch negatives.
Given a batch of $B$ training instances, the loss is defined as:
\begin{equation}
\label{loss:state}
    L_\textrm{state} = - \frac{1}{B} \sum_{i=1}^B \log \frac{e^{s(q_i, d_i^+)/\tau}}{Z_i},
\end{equation}
where $s(\cdot, \cdot)$ denotes the cosine similarity between the embeddings produced by the pooling layer and nonlinear head for the query and document, and $\tau$ is the temperature parameter. 
The normalization term $Z_i = \sum_{j=1}^B e^{s(q_i, d_j)/\tau}$ aggregates the scores across the positive pair and all in-batch negatives.

\subsection{State-Based Reranking}\label{sec:method-rerank}
State-based reranking aims to efficiently utilize learned states for reranking.

\paragraph{Model Architecture.}
The state-based RWKV-Reranker consists of RWKV blocks followed by a ranking head, as illustrated in Fig.~\ref{fig:arch}(b).
It takes the matrix-valued state as input and outputs a relevance score.
There are two modes for using the state-based RWKV-Reranker: offline mode and online mode.

In offline mode, document states $S_d$ are precomputed and cached.
We initialize the EmbeddingRWKV with the cached document state $S_d$ and then feed only the query tokens $\{x_t^q\}_{t=1}^{T_q}$ into the \textbf{frozen} EmbeddingRWKV to obtain the updated state $S'$.
The updated state $S'$ is fed into the state-based reranker to produce the relevance score.

In online mode, the document and query are processed jointly by the same \textbf{frozen} EmbeddingRWKV to produce the state on the fly for reranking.
Unless otherwise specified, we freeze all RWKV blocks in EmbeddingRWKV and train only the parameters of the state-based RWKV-Reranker.

\paragraph{Reranking Training Objective.}
For the reranking task, we formulate the relevance prediction as a binary classification problem and optimize the model using the Binary Cross-Entropy (BCE) loss:
\begin{equation}
\label{loss:reranking}
    L_\textrm{reranking} = - \big[ y \log s + (1 - y) \log (1 - s) \big],
\end{equation}
where $y \in \{0, 1\}$ denotes the ground truth label. $s = s(q, d)$ represents the predicted relevance probability for the query-document pair, derived from the state-based reranker's output.

\paragraph{Computational Cost Analysis.}
Transformer-based rerankers rely on full attention between query and document tokens, resulting in a quadratic computational complexity.
This cost grows rapidly with document length and becomes a major bottleneck for long-document reranking.

In contrast, our state-based reranker caches document states and processes only query tokens at inference time.
This design decouples inference latency from document length and achieves linear-time computation with respect to the query length.

\paragraph{Memory Footprint Analysis.}
We compare the memory cost of caching representations using Transformer’s KV caches and our matrix-valued states.
The ratio between their memory footprints is given by
\[
\frac{\text{Bytes}_{\text{token}}}{\text{Bytes}_{\text{state}}}
= \frac{2 L H S T b}{L H S^{2} b}
= \frac{2T}{S}.
\]
Here, $L$ denotes the number of layers, $H$ the number of heads, $S$ the per-head state dimension, $T$ the sequence length, and $b$ the number of bytes per element.
For RWKV-7 with $S=64$, this ratio becomes $T/32$.
In other words, the KV cache of a Transformer requires $T/32$ times more memory than a single state.
While the Transformer memory footprint grows linearly with sequence length $T$, the RWKV state remains constant, making it significantly more memory-efficient for long-document reranking.

\section{Experiments}
\subsection{Retrieval Experiments and Results}
We trained our embedding models (EmbeddingRWKV) using contrastive learning on our self-curated open-source dataset (Details are in Appendix~\ref{sec:detail_data}). 
We report the performance of our models on the MTEB English benchmark~\cite{muennighoff-etal-2023-mteb}.

\paragraph{EmbeddingRWKV is competitive.}
Table~\ref{tab:mteb_detail_en_v2} shows that EmbeddingRWKV delivers strong retrieval quality across three model scales (0.1B, 0.4B, and 1.4B parameters), with performance improving monotonically as model capacity increases.
In the base-size regime, EmbeddingRWKV-0.1B achieves competitive mean scores on MTEB English, matching or surpassing widely adopted open-source baselines of similar scale.
At the medium scale, EmbeddingRWKV-0.4B further advances performance and remains competitive with strong Transformer-based embedding models in the same parameter range.
When scaled to 1.4B parameters, EmbeddingRWKV-1.4B reaches near-parity with top-performing large embedding models and commercial APIs on the overall MTEB mean, while exhibiting particularly strong results on classification-oriented subsets, suggesting that the learned state representations capture discriminative semantic structure effectively.
Overall, these results indicate that RWKV-based backbones can serve as a competitive alternative to Transformer models for dense retrieval, without sacrificing embedding quality.
\begin{table*}[ht]
  \centering
  \renewcommand\arraystretch{1.0}
  \tabcolsep=0.08cm
  \caption{Comparison of embedding models on MTEB (English, v2). $^\gamma$Taken from \citep{lee2025gemini}. For other compared models, the scores are retrieved from MTEB online \href{https://huggingface.co/spaces/mteb/leaderboard}{leaderboard}.}
  \footnotesize
  \resizebox{\textwidth}{!}{%
  \begin{tabular}{lcc|cccccc}
    \toprule
    \textbf{Model} & \textbf{Size} & \textbf{Mean} & \textbf{Classification} & \textbf{Clustering} & \textbf{Reranking} & \textbf{Retrieval} & \textbf{STS} & \textbf{Summ.} \\
    \hline
    e5-base-v2~\citep{e5-base-v2} & 109M & 62.67 & 75.48 & 45.20 & 45.13 & 49.67 & 80.64 & 34.26 \\
    e5-base~\citep{e5-base} & 109M & 61.79 & 75.20 & 43.77 & 44.86 & 47.70 & 80.70 & 33.38 \\
    gte-base~\citep{gte-base} & 109M & 63.90 & 75.04 & 47.74 & 47.17 & 51.90 & 82.17 & 30.90 \\
    jina-embedding-b-en-v1~\citep{jina-embedding-b-en-v1} & 110M & 59.76 & 72.06 & 39.95 & 46.98 & 46.38 & 80.00 & 27.15 \\
    granite-embedding-english-r2~\citep{granite-embedding-english-r2} & 149M & 62.84 & 70.71 & 47.20 & 49.10 & 56.43 & 78.12 & 29.31 \\
    granite-embedding-125m-english~\citep{granite-embedding-125m-english} & 125M & 62.08 & 68.29 & 47.18 & 49.35 & 55.65 & 77.56 & 29.34 \\
    nomic-embed-text-v1.5~\citep{nomic-embed-text-v1.5} & 137M & 62.20 & 75.71 & 47.55 & 46.01 & 47.97 & 78.70 & 28.56 \\
    snowflake-arctic-embed-m-v1.5~\citep{snowflake-arctic-embed-m-v1.5} & 109M & 61.51 & 70.71 & 44.65 & 45.90 & 58.05 & 72.96 & 29.89 \\
    mmlw-roberta-base~\citep{mmlw-roberta-base} & 124M & 61.15 & 77.67 & 46.26 & 46.69 & 40.15 & 81.81 & 30.83 \\
    \rowcolor{lightgray}EmbeddingRWKV-0.1B & 144M & 63.06 & 81.01 & 46.21 & 46.07 & 48.06 & 80.30 & 25.13 \\
    \hline
    SearchMap-Preview~\citep{SearchMap-Preview} & 435M & 64.08 & 74.97 & 48.47 & 47.90 & 52.21 & 81.56 & 33.25 \\
    bilingual-embedding-large~\citep{bilingual-embedding-large} & 559M & 63.77 & 77.17 & 46.53 & 46.25 & 46.86 & 86.00 & 32.95 \\
    mmlw-e5-large~\citep{mmlw-e5-large} & 559M & 62.32 & 79.63 & 48.30 & 47.63 & 41.37 & 81.35 & 34.07 \\
    gte-large~\citep{gte-large} & 335M & 64.77 & 75.47 & 48.20 & 47.84 & 53.29 & 83.27 & 32.90 \\
    snowflake-arctic-embed-l~\citep{snowflake-arctic-embed-l} & 335M & 62.04 & 69.59 & 45.70 & 44.71 & 59.04 & 75.42 & 20.38 \\
    mmlw-roberta-large~\citep{mmlw-roberta-large} & 434M & 61.80 & 79.66 & 47.89 & 47.56 & 39.69 & 81.20 & 34.97 \\
    e5-large-v2~\citep{e5-large-v2} & 335M & 62.79 & 76.44 & 45.23 & 45.72 & 49.31 & 80.67 & 32.34 \\
    \rowcolor{lightgray}EmbeddingRWKV-0.4B & 389M & 64.86 & 86.21 & 47.78 & 46.92 & 49.42 & 79.58 & 31.37 \\
    \hline
    sentence-croissant-alpha-v0.4~\citep{sentence-croissant-alpha-v0.4} & 1B & 57.71 & 70.34 & 43.22 & 44.65 & 43.18 & 75.34 & 28.46 \\
    gte-Qwen2-1.5B-instruct~\citep{gte-Qwen2-1.5B-instruct} & 1.5B & 67.20 & 85.84 & 53.54 & 49.25 & 50.25 & 82.51 & 33.94 \\
    $^\gamma$OpenAI Commercial APIs: text-embedding-3-large & - & 66.43 & - & - & - & - & - & - \\
    $^\gamma$Cohere Commercial APIs: cohere-embed-multilingual-v3.0 & - & 66.01 & - & - & - & - & - & - \\
    \rowcolor{lightgray}EmbeddingRWKV-1.4B & 1.4B & 66.41 & 87.52 & 48.16 & 48.22 & 52.43 & 81.04 & 32.73 \\
    \bottomrule
  \end{tabular}%
  }
  \vspace{-0.2cm}
  \label{tab:mteb_detail_en_v2}
\end{table*}

\paragraph{Curriculum training is data-efficient.}
Table~\ref{tab:curriculum_results} compares our single-stage domain-aware curriculum training with a conventional multi-stage pipeline (pretraining followed by SFT).
The multi-stage baseline uses 132.1M training samples in total, while our curriculum uses only 6.7M samples (about 5\%).
Despite this large reduction in data and a simpler training recipe, the curriculum-trained EmbeddingRWKV achieves higher MTEB English performance than the multi-stage baseline, i.e., Curriculum (6.7M) $>$ Multi-stage (132.1M).
These results indicate that curriculum training provides a stronger and more data-efficient way to learn reusable state representations, which not only improves retrieval but also supplies high-quality states for the subsequent state-based reranking stage.
\begin{table}[h]
  \centering
  \renewcommand\arraystretch{1.1} 
  \tabcolsep=0.12cm
  \caption{Comparison of retrieval performance between standard Multi-stage training and Single-stage Domain-Aware Curriculum training. The curriculum method achieves higher scores using only $\sim$5\% of the baseline's total training data.}
  \footnotesize
  \resizebox{\linewidth}{!}{%
  \begin{tabular}{lcc|c|cc}
    \toprule
    \textbf{Model} & \textbf{Stage} & \textbf{Size} & \textbf{Sample Size} & \textbf{MTEB-E} & \textbf{MTEB-C} \\
    \midrule
    \rowcolor{lightgray}\multicolumn{6}{l}{\textit{Multi-stage Training}} \\
    EmbeddingRWKV & Pretrain & 144M & 123.2M & 53.85 & 51.47 \\
    EmbeddingRWKV & SFT & 144M & 8.9M & 59.60 & 54.37 \\
    EmbeddingRWKV & Pretrain & 389M & 123.2M & 54.88 & 53.69 \\
    EmbeddingRWKV & SFT & 389M & 8.9M & 60.85 & 55.17 \\
    \midrule
    \rowcolor{lightgray}\multicolumn{6}{l}{\textit{Single-stage Curriculum Training}} \\
    EmbeddingRWKV & SFT & 144M & 6.7M & \textbf{63.06} & \textbf{57.12} \\
    EmbeddingRWKV & SFT & 389M & 6.7M & \textbf{64.86} & \textbf{58.68} \\
    \bottomrule
  \end{tabular}%
  }
  \vspace{-0.2cm}
  \label{tab:curriculum_results}
\end{table}

\subsection{Reranking Experiments and Results}
We freeze EmbeddingRWKV and train the state-based RWKV-Reranker with supervised learning. We report reranking results on NanoBEIR, MTEB (English, v2) Retrieval, and MTEB (Chinese, v1) Retrieval.

\paragraph{State representation learning is necessary.}
A natural question is whether we can directly reuse states from a base RWKV language model for reranking, without retrieval-oriented state representation learning.
Table~\ref{tab:base_vs_emb} shows that this is not the case: initializing the reranker with states from the base LM consistently underperforms using states from EmbeddingRWKV across all evaluated metrics and benchmarks.
We attribute this gap to the mismatch in training objectives.
Reranking requires fine-grained relevance matching and a retrieval-specific representation geometry, whereas base LMs primarily learn generative language modeling signals that are not aligned with retrieval supervision.
Therefore, explicitly learning retrieval-aligned state representations is crucial for effective state-based reranking.
\begin{table}[h]
\centering
\caption{Comparison of state-based reranking performance using states generated by RWKV-7 LM and EmbeddingRWKV on NanoBEIR.}
\label{tab:base_vs_emb}
\resizebox{\columnwidth}{!}{
\begin{tabular}{lcccc}
\toprule
\textbf{State Backbone} & \textbf{Size} & \textbf{MAP} & \textbf{MRR@10} & \textbf{NDCG@10} \\
\midrule
RWKV-7 LM & 190M & 48.58 & 59.37 & 54.78 \\
EmbeddingRWKV & 144M & 58.29 & 69.50 & 63.41 \\
\bottomrule
\end{tabular}
} 
\end{table}

\paragraph{State-based reranking is effective.}
A central question in our design is whether the state-based mechanism compromises ranking quality compared to traditional architectures. To investigate this, we trained a standard RWKV-based single-tower reranker under an identical configuration (0.1B parameters) to our proposed model. As shown in Table~\ref{tab:reranker_results_nano_mteb}, the state-based reranker demonstrates that our model achieves performance parity with single-tower baseline on NanoBEIR, MTEB-E (eng, v2), and MTEB-C (cmn, v1). Furthermore, scaling the model from 90M to 1.3B parameters yields monotonic quality improvements, validating the favorable scaling behavior of the state-based reranker.

We additionally compare against three Transformer-based rerankers, including ModernBERT~\citep{Warner2024SmarterBF}, EuroBERT~\citep{Boizard2025EuroBERTSM}, and the Jina reranker~\cite{Wang2025jinarerankerv3LB}. Overall, our state-based reranker exhibits only a small gap to strong single-tower Transformer rerankers, while maintaining performance parity with these Transformer baselines on the NanoBEIR benchmark.
\begin{table}[h]
  \centering
  \renewcommand\arraystretch{1.0}
  \tabcolsep=0.1cm
  \caption{State-Based Reranker performance on NanoBEIR~(NDCG@10) and MTEB-E (English, v2) Retrieval, MTEB-C (Chinese, v1) Retrieval. Scores are scaled by 100 (higher is better).}
  \footnotesize
  \resizebox{\linewidth}{!}{%
  \begin{tabular}{lc|c|c|c}
    \toprule
    \textbf{Model} & \textbf{Size} & \textbf{NanoBEIR} & \textbf{MTEB-E-Retr} & \textbf{MTEB-C-Retr} \\
    \midrule
    EmbeddingRWKV & 0.1B & 59.10 & 48.06 & 57.26 \\
    \midrule
    \rowcolor{lightgray}\multicolumn{5}{l}{\textit{Traditional Single-Tower Reranker}} \\
    RWKV-Reranker & 90M & 64.88 & 49.31 & 58.16 \\
    EuroBERT & 210M & 47.38 & - & - \\
    ModernBERT & 395M & 64.18 & - & - \\
    Jina Reranker & 278M & 70.68 & - & - \\
    \midrule
    \rowcolor{lightgray}\multicolumn{5}{l}{\textit{State-Based Reranker}} \\
    RWKV-Reranker & 90M & 63.41 & 47.78 & 58.42 \\
    RWKV-Reranker & 317M & 68.60 & 53.22 & 63.60 \\
    RWKV-Reranker & 1.3B & 71.58 & 56.44 & 66.30 \\
    \bottomrule
  \end{tabular}%
  }
  \vspace{-0.2cm}
  \label{tab:reranker_results_nano_mteb}
\end{table}

\begin{figure}[htbp]
    \centering
    \includegraphics[width=\linewidth]{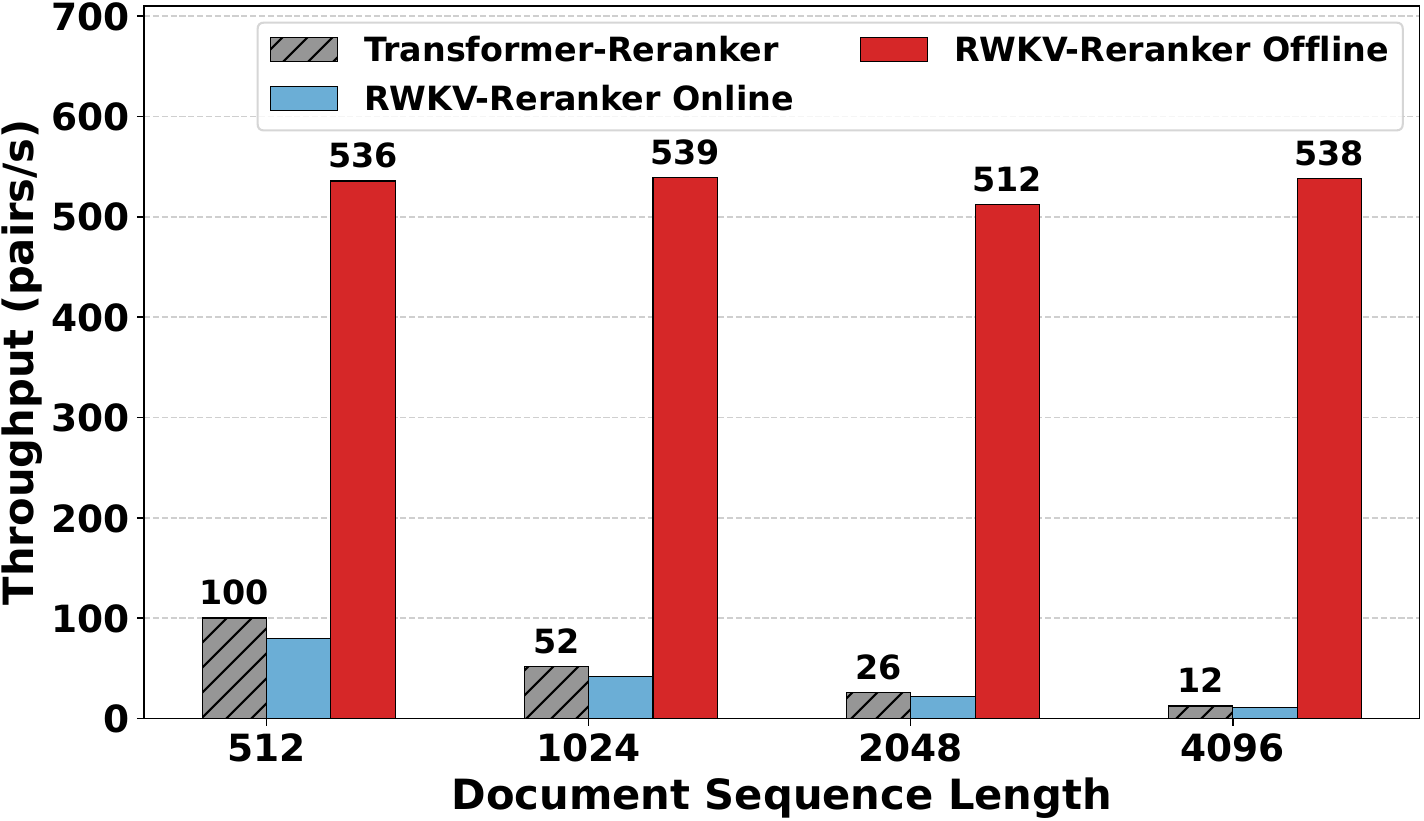}
    \includegraphics[width=\linewidth]{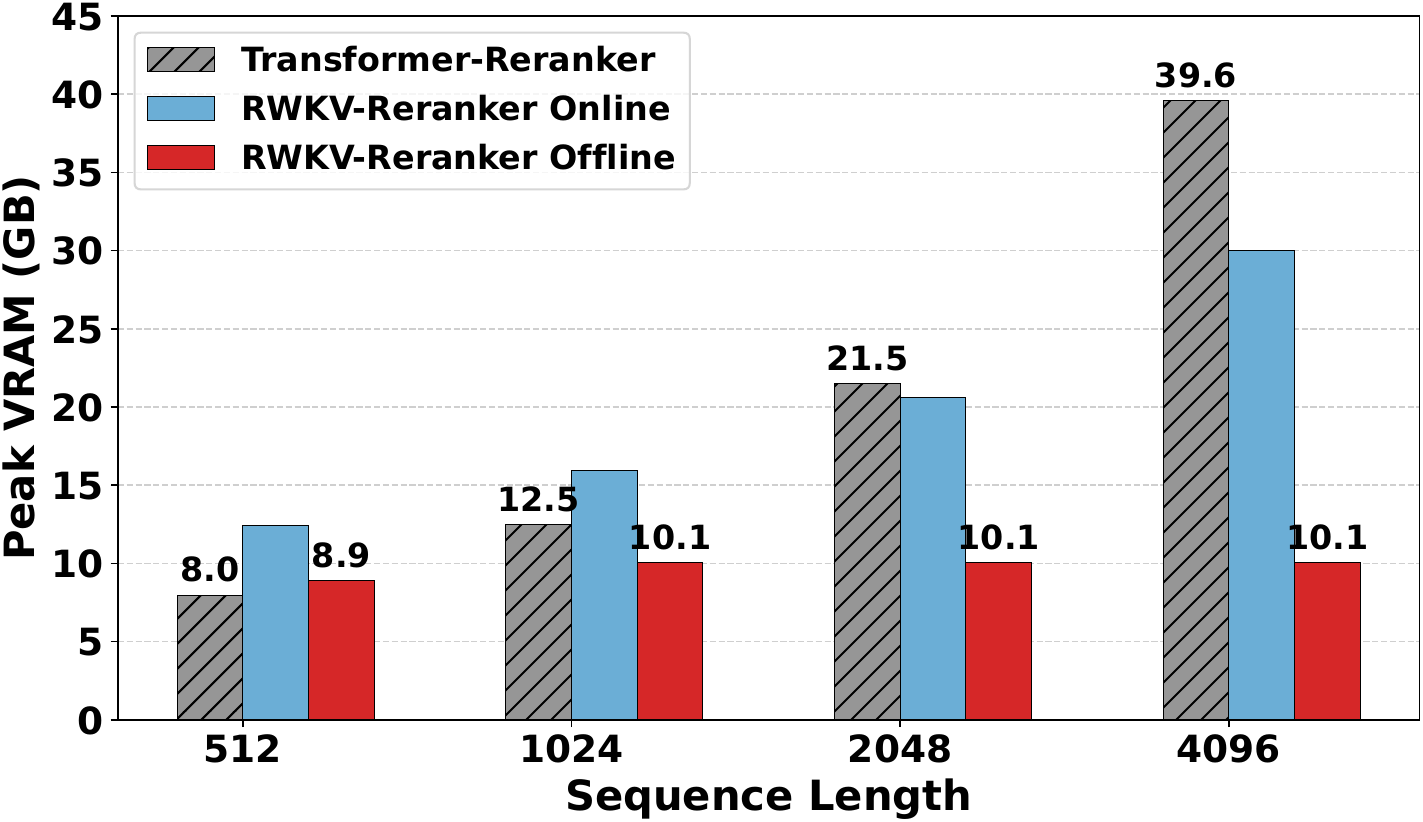}
    \caption{Performance benchmarking across various sequence lengths. Baseline: mxbai-rerank-large-v2 (1.5B)~\cite{li2025prorank} with FlashAttention-2 vs. Ours: EmbeddingRWKV-1.4B + RWKV-Reranker-1.3B.
    Top: Throughput comparison (pair/s), where higher is better.
    Bottom: Peak VRAM usage (GB), where lower is better.}
    \label{fig:throughput_and_vram_comparison}
\end{figure}

\paragraph{State-based reranking is efficient.}
As shown in Figure~\ref{fig:throughput_and_vram_comparison}, our offline state-based reranker delivers substantial acceleration over a strong Transformer reranker with FlashAttention-2~\citep{dao2023flashattention,li2025prorank}, with speedup increasing from 5.4$\times$ at a document length of 512 to 44.8$\times$ at 4096, while also using markedly less GPU memory.
The key mechanism is that offline reranking processes only query tokens together with cached document states; thus inference cost is independent of document length.
This decoupling is reflected by the scaling curves: Transformer-based rerankers exhibit rapidly degrading throughput as document length increases, whereas the RWKV offline reranker remains nearly flat across lengths.
Similarly, peak VRAM usage for attention-based baselines grows sharply with longer documents, while our approach stays stable due to constant-size state.
Moreover, even in the online setting, the state-based reranker is more memory-efficient than Transformer-based rerankers: its peak VRAM scales more slowly with sequence length, consistent with RWKV’s linear memory scaling.
Additional efficiency breakdowns across model sizes are provided in Appendix~\ref{sec:detail_reranker} and Table~\ref{tab:rwkv_reranker_benchmark}.

\paragraph{State is redundant.}
An important practical question is how many states to cache and which layers to keep, i.e., where the compression boundary lies.
Table~\ref{tab:reranker_layer_selection} shows substantial redundancy across layers: keeping only a small subset of layer states preserves most of the full-depth performance.
For example, using 50\% of the layers already retains 99.59\% of the performance, and even more aggressive configurations with only a few uniformly selected layers can still achieve 97.70\%--98.62\% relative performance.
These results suggest that reranking signals are highly compressible, with overlapping information distributed across model depth rather than concentrated in a single layer.

Importantly, we find that layer choice matters more than layer count.
With the same three-layer budget, a top-heavy selection leads to a sharp drop (85.99\%), whereas a uniformly distributed selection recovers performance to 98.62\%.
This indicates that effective reranking relies on combining multi-level signals spanning early to late layers.
Based on this observation, we adopt uniform stride layer selection as the default strategy in our state caching design.
\begin{table*}[ht]
\centering
\caption{Ablation on state-based reranker layer selection strategies. \textbf{Mean} is the average of NanoBEIR, MTEB-E-Retr, and MTEB-C-Retr scores. \textbf{Ratio} indicates performance retention relative to the best configuration in the same model group. \textbf{Uniform} selection strategies (e.g., 1, 6, 10) significantly outperform contiguous top-layer strategies.}
\label{tab:reranker_layer_selection}
\resizebox{\linewidth}{!}{
\begin{tabular}{lcccccccc}
\toprule
    \textbf{Strategy} & \textbf{Layers} & \textbf{Layer Index} & \textbf{Size} & \textbf{NanoBEIR} & \textbf{MTEB-E-Retr} & \textbf{MTEB-C-Retr} & \textbf{Mean} & \textbf{Ratio} \\
\midrule
\rowcolor{lightgray}\multicolumn{9}{l}{\textit{RWKV-Reranker-0.1B (Full Depth: 12 Layers)}} \\
\quad Top-heavy &               3 &                 9-11 &         23.1M &             56.94 &           38.86 &           50.07 &         48.62 &        85.99\% \\
\quad Uniform &               3 &             0, 5, 11 &         23.1M &             61.85 &           44.82 &           55.71 &         54.13 &        95.74\% \\
\quad Uniform &               3 &             1, 6, 10 &         23.1M &             62.35 &           48.05 &           56.88 &         55.76 &        98.62\% \\
\quad Top-heavy &               6 &                 6-11 &         45.8M &             62.01 &           46.99 &           57.65 &         55.55 &        98.25\% \\
\quad Uniform &               6 &    0, 3, 5, 7, 9, 11 &         45.8M &             64.08 &           47.52 &           57.32 &         56.31 &        99.59\% \\
\quad Full &              12 &                 0-11 &         90.0M &             63.41 &           47.78 &           58.42 &         56.54 &        100.0\% \\
\midrule
\rowcolor{lightgray}\multicolumn{9}{l}{\textit{RWKV-Reranker-1.3B (Full Depth: 24 Layers)}} \\
\quad Top-heavy &               1 &                   23 &           54M &             50.02 &           35.55 &           45.10 &         43.56 &        67.25\% \\
\quad Uniform &               3 &              1,11,22 &          161M &             70.33 &           54.92 &           64.58 &         63.28 &         97.70\% \\
\quad Top-heavy &               6 &                18-23 &          318M &             69.01 &           51.43 &           64.39 &         61.61 &        95.12\% \\
\quad Uniform &               6 &      1,6,11,15,19,22 &          318M &             71.19 &           55.10 &           65.51 &         63.93 &         98.70\% \\
\quad Full &              24 &                 0-23 &          1.3B &             71.58 &           56.44 &           66.30 &         64.77 &        100.0\% \\
\bottomrule
\end{tabular}
}
\end{table*}

\paragraph{Sharing state fails catastrophically.}
Although layer pruning suggests substantial redundancy, compression is not arbitrary.
As a counterexample, Table~\ref{tab:share_state_ablation} shows that forcing all layers to share a single state from the final layer leads to a marked performance degradation.
This result clarifies that redundancy does not imply that all useful signals can be collapsed into one state.

We attribute this failure to the functional specialization across depth.
Intermediate-layer states tend to preserve fine-grained matching cues and local evidence, while the final-layer state is more abstract and task-agnostic.
Hard sharing therefore removes the multi-scale signals required for accurate relevance estimation, resulting in catastrophic degradation in reranking quality.
\begin{table}[ht]
\centering
\caption{Ablation on sharing a single state across layers in RWKV-Reranker 0.1B.}
\label{tab:share_state_ablation}
\resizebox{\columnwidth}{!}{
\begin{tabular}{lcccc}
\toprule
\textbf{States} & \textbf{Layers} & \textbf{NanoBEIR} & \textbf{MTEB-E-Retr} & \textbf{MTEB-C-Retr}\\
\midrule
All states & 12 & 63.41 & 47.78 & 58.42 \\
One shared state & 12 & 51.23 & 11.00 & 5.62 \\
\bottomrule
\end{tabular}
}
\end{table}

\paragraph{Why not cache Transformer KV for reranking?}
A natural question is whether a similar idea can be applied to Transformers: can we cache document KV caches to accelerate reranking?
The answer is no, for a simple reason---the KV cache is prohibitively large.
For a document of 2000 tokens, a Transformer reranker with comparable parameter scale would require \emph{62.5$\times$} more storage to cache its KV states than our matrix-valued state.
This makes KV-based caching impractical at scale, especially when reranking requires caching states for large corpora.
In contrast, RWKV-style models maintain a constant-size matrix-valued state, which is exactly the architectural property we aim to exploit: we design the reranking pipeline around this compact, reusable representation to achieve efficient inference.
A detailed discussion are provided in Appendix~\ref{sec:per_doc_cache}.

\section{Conclusion}

We propose a state-centric retrieval paradigm, leveraging states as a bridge to achieve architectural synergy between embedding and reranking stages.
This approach enables rerankers to reuse cached states, achieving high-quality results with improved efficiency.
Beyond retrieval, our results suggest that direct modeling over states is both feasible and promising. 
This paradigm opens up opportunities for extending state-centric modeling to a broader range of scenarios, including state-based value models, verifiers, and reward models. 
We believe that state-centric modeling has significant potential to generalize to diverse application scenarios, particularly in the development of intelligent agents, where it may be naturally integrated into traditional reinforcement learning frameworks.

\section*{Limitations}
Despite its strong empirical performance and efficiency benefits, our work has several limitations that warrant further investigation.
Although our state-based reranking reduces the dependence on document length at inference time, the overall pipeline still requires running and maintaining two models. This introduces additional resource.
Our reranker operates on cached matrix-valued states, which are a compressed summary of the document. For extremely long or highly detailed documents, this compression may discard fine-grained evidence, potentially leading to accuracy degradation.

\section*{Acknowledgments}
We thank Yuanshi Intelligence Co. for providing the computational resources and infrastructure support that made this research possible.

\bibliography{custom}

\appendix
\newpage
\onecolumn
\section{Implementation Details}
\paragraph{Evaluation Setup}
We evaluate our method on MTEB (eng, v2) for embeddings, and on NanoBEIR, MTEB (eng, v2), and MTEB (cmn, v1) for reranking tasks.
We adopt a standard two-stage retrieval pipeline, utilizing an initial retriever to generate candidates followed by a reranker for fine-grained sorting, ensuring an optimal trade-off between retrieval latency and precision.
Specifically, we employ a 0.1B EmbeddingRWKV model to retrieve the top-100 document candidates, which are subsequently re-ranked by the reranker.

\paragraph{Hyperparameters}
We show the training hyperparameters for both EmbeddingRWKV models and RWKV-Reranker models in Table~\ref{tab:hyperparameter}. 

\begin{table}[h!]
\centering
\resizebox{\textwidth}{!}{
\begin{tabular}{l|c ccccc}
\toprule
Hyperparameter & EmbeddingRWKV-0.1B& EmbeddingRWKV-0.4B & EmbeddingRWKV-1.4B& RWKV-Reranker-0.1B& RWKV-Reranker-0.3B&RWKV-Reranker-1.3B\\
\midrule
batch size & 1024 & 1024  & 512 & 512  & 512 &512  \\
lr init& 1e-5 & 1e-5& 1e-5 & 1e-5& 1e-5 &1e-5\\
lr end& 1e-5 & 1e-5& 1e-5 & 5e-6& 5e-6 &5e-6\\
lr schedule & constant&constant&constant&cosine decay&cosine decay&cosine decay\\
lr warmup ratio & 0&0&0&0&0&0\\
weight decay & 0&0&0&0&0&0\\
epoch & 1& 1&  1& 1& 1&1\\
optimizer & AdamW&AdamW&AdamW&AdamW&AdamW&AdamW\\
DeepSpeed stage & 1& 1 & 1& 1& 1&1\\
\bottomrule
\end{tabular}
}
\caption{
\textbf{Hyperparameters} of EmbeddingRWKV and RWKV-Reranker.
}
\label{tab:hyperparameter}
\end{table}

\section{Additional Results and Discussions}
\label{sec:add_exp_results}
\subsection{Efficiency Analysis of State-Based Reranking}
\label{sec:detail_reranker}
We analyze Table~\ref{tab:rwkv_reranker_benchmark}, comparing RWKV-Reranker (Online/Offline) with attention-based baselines across Large ($\approx$1.5B), Base ($\approx$0.4B), and Small ($\approx$0.1B) model regimes.

\paragraph{Throughput and sequence-length scaling}
The key advantage appears in RWKV Offline, where document states/embeddings are pre-computed and reused.
\begin{itemize}
\item \textbf{Large models (1.5B):} The attention baseline throughput degrades sharply with length (97.3 pairs/s at 512 $\rightarrow$ 11.9 pairs/s at 4096), while RWKV Offline stays nearly constant at \textbf{$\sim$500+ pairs/s} across lengths.
\item \textbf{Small models (0.1B):} FlashAttention remains fast at short contexts but still slows down markedly at long contexts. In contrast, RWKV Offline sustains \textbf{$\sim$2440 pairs/s} up to 4096 tokens, reaching a \textbf{75.6$\times$} speedup at the longest setting.
\end{itemize}

\paragraph{Memory efficiency (VRAM)}
RWKV’s linear-state computation avoids the quadratic activation/memory growth of attention. At 4096 tokens, Standard Attention peaks at \textbf{39.59 GB} VRAM, whereas RWKV Offline requires only \textbf{10.07 GB}, i.e., an \textbf{$\approx$4$\times$} reduction. This substantially lowers deployment barriers and enables larger inference batch sizes.

\paragraph{Online vs. Offline modes}
RWKV Online processes the full sequence on-the-fly and is therefore bounded by serial state updates; it primarily improves memory scaling rather than throughput. The intended operating point of our design is \textbf{split processing (Offline)}, which decouples document encoding from query interaction and reduces reranking to the lightweight query-stage computation (e.g., \textbf{$\sim$38 ms} for large models).
\begin{table}[H]
    \centering
    \caption{Performance comparison of Attention mechanisms vs RWKV-Reranker (Online/Offline) across various sequence lengths (Query Len = 64, Batch Size = 100).}
    \label{tab:rwkv_reranker_benchmark}
    \resizebox{\linewidth}{!}{%
        \begin{tabular}{l c c c c c c c}
            \toprule
            \textbf{Mode} & \textbf{Doc Len} & \textbf{Total (ms)} & \textbf{Embed (ms)} & \textbf{Rerank (ms)} & \textbf{Peak VRAM} & \textbf{Throughput (pair/s)} & \textbf{Speedup} \\
            \midrule
            \rowcolor{lightgray}\multicolumn{8}{l}{\textit{ Baseline: mxbai-rerank-large-v2 (1.5B) \quad vs. \quad Ours: EmbeddingRWKV-1.4B + RWKV-Reranker-1.3B}} \\
            Standard Attention & 512  & 1027.48 & -     & -   & 7.97 GB  & 97.3  & 1.0x \\
            Flash Attention    & 512  & 997.66  & -     & -   & 7.97 GB  & 100.2 & 1.0x \\
            RWKV-Reranker Online        & 512  & 1247.90 & 1209.66 & 38.21 & 12.41 GB & 80.1  & 0.8x \\
            RWKV-Reranker Offline       & 512  & 186.70  & 148.67  & 38.02 & 8.88 GB  & 535.6 & 5.5x \\
            \midrule
            Standard Attention & 1024 & 1988.10 & -     & -   & 12.49 GB & 50.3  & 1.0x \\
            Flash Attention    & 1024 & 1926.90 & -     & -   & 12.49 GB & 51.9  & 1.0x \\
            RWKV-Reranker Online        & 1024 & 2394.71 & 2356.14 & 38.55 & 15.95 GB & 41.8  & 0.8x \\
            RWKV-Reranker Offline       & 1024 & 185.46  & 148.50  & 36.94 & 10.07 GB & 539.2 & 10.7x \\
            \midrule
            Standard Attention & 2048 & 4026.59 & -     & -   & 21.52 GB & 24.8  & 1.0x \\
            Flash Attention    & 2048 & 3860.74 & -     & -   & 21.52 GB & 25.9  & 1.0x \\
            RWKV-Reranker Online        & 2048 & 4598.01 & 4560.24 & 37.74 & 20.63 GB & 21.7  & 0.9x \\
            RWKV-Reranker Offline       & 2048 & 195.35  & 147.95  & 47.37 & 10.07 GB & 511.9 & 20.6x \\
            \midrule
            Standard Attention & 4096 & 8410.25 & -     & -   & 39.59 GB & 11.9  & 1.0x \\
            Flash Attention    & 4096 & 8024.71 & -     & -   & 39.59 GB & 12.5  & 1.0x \\
            RWKV-Reranker Online        & 4096 & 9003.55 & 8965.33 & 38.20 & 30.01 GB & 11.1  & 0.9x \\
            RWKV-Reranker Offline       & 4096 & 185.99  & 147.52  & 38.45 & 10.07 GB & 537.7 & 45.2x \\
            \midrule
            \rowcolor{lightgray}\multicolumn{8}{l}{\textit{ Baseline: mxbai-rerank-base-v2 (0.5B) \quad vs. \quad Ours: EmbeddingRWKV-0.4B + RWKV-Reranker-0.3B}} \\
            Standard Attention & 512  & 381.15  & -      & -     & 3.55 GB  & 262.4  & 1.0x \\
            Flash Attention    & 512  & 360.84  & -      & -     & 3.55 GB  & 277.1  & 1.1x \\
            RWKV-Reranker Online        & 512  & 479.16  & 444.06 & 35.08 & 5.02 GB  & 208.7  & 0.8x \\
            RWKV-Reranker Offline       & 512  & 96.07   & 59.55  & 36.50 & 3.26 GB  & 1040.9 & 4.0x \\
            \midrule
            Standard Attention & 1024 & 723.18  & -      & -     & 5.87 GB  & 138.3  & 1.0x \\
            Flash Attention    & 1024 & 687.43  & -      & -     & 5.87 GB  & 145.5  & 1.1x \\
            RWKV-Reranker Online        & 1024 & 913.71  & 878.26 & 35.43 & 6.79 GB  & 109.4  & 0.8x \\
            RWKV-Reranker Offline       & 1024 & 94.10   & 59.66  & 34.43 & 3.85 GB  & 1062.7 & 7.7x \\
            \midrule
            Standard Attention & 2048 & 1461.67 & -      & -     & 10.51 GB & 68.4   & 1.0x \\
            Flash Attention    & 2048 & 1385.56 & -      & -     & 10.51 GB & 72.2   & 1.1x \\
            RWKV-Reranker Online        & 2048 & 1747.95 & 1712.42& 35.51 & 9.14 GB  & 57.2   & 0.8x \\
            RWKV-Reranker Offline       & 2048 & 94.53   & 59.88  & 34.64 & 3.85 GB  & 1057.8 & 15.5x \\
            \midrule
            Standard Attention & 4096 & 3098.93 & -      & -     & 19.79 GB & 32.3   & 1.0x \\
            Flash Attention    & 4096 & 2946.56 & -      & -     & 19.79 GB & 33.9   & 1.1x \\
            RWKV-Reranker Online        & 4096 & 3424.99 & 3388.77& 36.19 & 13.82 GB & 29.2   & 0.9x \\
            RWKV-Reranker Offline       & 4096 & 95.42   & 60.05  & 35.34 & 3.85 GB  & 1048.0 & 32.5x \\
            \midrule
            \rowcolor{lightgray}\multicolumn{8}{l}{\textit{ Baseline: gte-reranker-modernbert-base (0.1B) \quad vs. \quad Ours: EmbeddingRWKV-0.1B + RWKV-Reranker-0.1B}} \\
            Standard Attention & 512  & 171.85  & -       & -     & 1.16 GB  & 581.9  & 1.0x \\
            Flash Attention    & 512  & 110.02  & -       & -     & 1.00 GB  & 908.9  & 1.6x \\
            RWKV-Reranker Online        & 512  & 178.52  & 160.77  & 17.74 & 2.34 GB  & 560.2  & 1.0x \\
            RWKV-Reranker Offline       & 512  & 41.00   & 23.34   & 17.65 & 1.24 GB  & 2438.9 & 4.2x \\
            \midrule
            Standard Attention & 1024 & 387.41  & -       & -     & 2.14 GB  & 258.1  & 1.0x \\
            Flash Attention    & 1024 & 212.34  & -       & -     & 1.62 GB  & 470.9  & 1.8x \\
            RWKV-Reranker Online        & 1024 & 333.24  & 315.05  & 18.18 & 3.45 GB  & 300.1  & 1.2x \\
            RWKV-Reranker Offline       & 1024 & 40.98   & 23.41   & 17.56 & 1.46 GB  & 2440.0 & 9.5x \\
            \midrule
            Standard Attention & 2048 & 1009.89 & -       & -     & 4.68 GB  & 99.0   & 1.0x \\
            Flash Attention    & 2048 & 443.68  & -       & -     & 2.87 GB  & 225.4  & 2.3x \\
            RWKV-Reranker Online        & 2048 & 620.39  & 602.03  & 18.33 & 5.21 GB  & 161.2  & 1.6x \\
            RWKV-Reranker Offline       & 2048 & 40.83   & 23.37   & 17.45 & 1.46 GB  & 2449.3 & 24.7x \\
            \midrule
            Standard Attention & 4096 & 3098.45 & -       & -     & 12.11 GB & 32.3   & 1.0x \\
            Flash Attention    & 4096 & 991.55  & -       & -     & 5.38 GB  & 100.9  & 3.1x \\
            RWKV-Reranker Online        & 4096 & 1214.92 & 1196.54 & 18.36 & 8.73 GB  & 82.3   & 2.6x \\
            RWKV-Reranker Offline       & 4096 & 40.99   & 23.42   & 17.55 & 1.46 GB  & 2439.8 & 75.6x \\
            \bottomrule
        \end{tabular}%
    }
\end{table}

\subsection{Computational Complexity of Backbone}

Unlike Transformer-based embedding models (e.g., BERT, GPT) which suffer from quadratic time complexity $O(L^2)$ due to the global self-attention mechanism, EmbeddingRWKV adopts RWKV as its backbone, which employs a linear attention formulation. This results in a time complexity of $O(L)$ and a constant memory footprint $O(1)$ regarding the sequence state during inference, irrespective of the input sequence length. 

As shown in Table~\ref{tab:complexity_comparison}, this property allows our model to scale efficiently to long-context tasks where traditional transformers typically encounter memory bottlenecks.

\begin{table}[htbp]
    \centering
    \caption{Computational complexity comparison between Transformer-based backbones and RWKV during inference. $L$ denotes the sequence length, and $d$ denotes the hidden dimension.}
    \label{tab:complexity_comparison}
    \vspace{0.2cm}
    \begin{tabular}{lcccc}
        \toprule
        \textbf{Architecture} & \textbf{Attention Mechanism} & \textbf{Time Complexity} & \textbf{Space Complexity} & \textbf{Max Context} \\
        \midrule
        BERT / GPT & $O(L^2)$ (Global) & $O(L^2)$ & $O(L^2 + Ld)$ & Finite \\
        FlashAttention* & $O(L^2)$ (Tiled) & $O(L^2)$ & $O(L)$ & Finite \\
        \textbf{RWKV} & \textbf{$O(L)$ (Linear)} & $\mathbf{O(L)}$ & $\mathbf{O(1)}^\dagger$ & \textbf{Infinite} \\
        \bottomrule
    \end{tabular}
    \vspace{0.1cm}
    \footnotesize
    \begin{flushleft}
        $^\dagger$ \textit{Note: RWKV maintains a fixed-size hidden state (RNN-mode) regardless of history length $L$, whereas Transformers must store the Attention Matrix or KV Cache.}
    \end{flushleft}
\end{table}

\subsection{Memory Footprint Usage}
\paragraph{Analytical Memory Cost}
We compare the per-document cache footprint of RWKV document states against Transformer KV caches under the same precision setting (FP16). 
For RWKV, each layer stores a document state with shape $(H, S, S)$ (one state per head), hence the cache size is 
$\text{Bytes}_{\text{RWKV}} = L \times H \times S^2 \times 2$, where $L$ is the number of layers, $H$ the number of heads, $S$ the state size, and $2$ accounts for FP16 bytes.

For Transformers, the document-side cache is the concatenated Key/Value tensors. Per layer it stores $(K,V)$ with total width $2d_{\text{model}}$ over $T$ cached tokens, giving 
$\text{Bytes}_{\text{KV}} = L \times (2d_{\text{model}}) \times T \times 2$.
To make the comparison interpretable, we use the common RWKV state--KV equivalence $T_{\text{eq}} = S/2$. 
With $S=64$, this yields $T_{\text{eq}}=32$, meaning one RWKV document state per layer is roughly comparable to caching $\sim32$ Transformer tokens (KV) per layer.

\paragraph{Per-document Cache Usage Comparison}
\label{sec:per_doc_cache}
Table~\ref{tab:cache_mb} reports the per-document cache size for RWKV states (with $S=64$) and Transformer KV caches at $T=2000$.
A key observation is that the ratio stays constant at $2000/32 = 62.5\times$ across model sizes: when $S$ and $T$ are fixed, both RWKV and KV scale linearly with $L$, while the dominant difference comes from RWKV being \emph{token-length independent} (state cache does not grow with $T$), whereas Transformer KV grows linearly with $T$.
Concretely, RWKV requires only 1--16 MB per document in the configurations shown, while Transformer KV reaches 70--1000 MB at $T=2000$.
This gap translates directly into higher cache density (more documents stored per GPU/CPU memory budget) and substantially lower retrieval-system memory pressure when large-scale document caching is required.

\begin{table}[h]
\centering
\small
\caption{Per-document cache usage for RWKV states vs. Transformer KV cache at $T=2000$ (FP16). RWKV uses $S=64$ and $T_{\text{eq}}=S/2=32$, leading to a constant ratio $T/T_{\text{eq}}=62.5\times$ under fixed $S$ and $T$.}
\label{tab:cache_mb}
\begin{tabular}{@{}lccccccc@{}}
\toprule
\textbf{Model Size} & \textbf{L} & \textbf{$\mathbf{d_{model}}$} & \textbf{H} & \textbf{S} & \textbf{\shortstack{RWKV State\\(MB)}} & \textbf{\shortstack{Transformer KV\\@ T=2000 (MB)}} & \textbf{\shortstack{Ratio\\(Trans/RWKV)}} \\ \midrule
0.1B & 12 & 768 & 12 & 64 & 1.12 & 70.31 & 62.5$\times$ \\
0.4B & 24 & 1024 & 16 & 64 & 3.00 & 187.50 & 62.5$\times$ \\
1.5B & 24 & 2048 & 32 & 64 & 6.00 & 375.00 & 62.5$\times$ \\
$\sim$3B & 32 & 2560 & 40 & 64 & 10.00 & 625.00 & 62.5$\times$ \\
$\sim$7B & 32 & 4096 & 64 & 64 & 16.00 & 1000.00 & 62.5$\times$ \\ \bottomrule
\end{tabular}
\end{table}

\section{Data Details}
\label{sec:detail_data}
\subsection{Pre-training Data Construction}
\label{sec:pretraining_data}

To build a robust and versatile embedding model, we curated a large-scale pre-training mixture of approximately 470M text pairs. The collection strategy targets (i) linguistic diversity, (ii) broad domain coverage, and (iii) multiple levels of text granularity. As summarized in Table~\ref{tab:pretrain_data_list}, the pre-training sources fall into three coarse groups: multilingual corpora, Chinese-focused corpora, and English (including code) resources.

Multilingual corpora provide the backbone for cross-lingual semantic alignment. We include large unstructured resources such as \textit{CC-News} (100M pairs) and \textit{Wikipedia} (100M pairs) to capture general knowledge and multilingual semantics. To strengthen instruction following and IR-oriented matching, we incorporate curated query--document style datasets including \textit{xP3}~\citep{DBLP:conf/acl/MuennighoffWSRB23} and the \textit{SWIM-IR} suite~\citep{DBLP:conf/naacl/ThakurNAWLC24}. We further leverage parallel corpora such as \textit{NLLB} to anchor representations across distant language families.

Given the unique challenges of Chinese semantic modeling, we explicitly up-weight high-quality Chinese sources. \textit{Wudao}~\citep{DBLP:journals/aiopen/YuanZDDLCZYT21} serves as a primary component (44M pairs) for high-coverage web text, complemented by vertical-domain corpora such as \textit{CSL} (scientific literature), \textit{THUCNews} (news), and \textit{Zhihu-KOL} (community Q\&A). This combination improves robustness across both formal registers and colloquial discourse.

English and code data are included to reinforce general semantic reasoning and technical coverage. We use \textit{Reddit} (100M pairs) and \textit{StackExchange} (14M pairs) for conversational and QA-style semantics, together with \textit{S2ORC} (41M pairs) and \textit{CodeSearchNet} for scientific and programming-oriented content. We additionally include \textit{PAQ}~\citep{DBLP:journals/tacl/LewisWLMKPSR21} to improve retrieval behavior via probable questions derived from Wikipedia.

Overall, this mixture is designed to be complementary along three axes. First, it spans diverse \textit{granularity}, from short texts (e.g., reviews, tweets) to long documents (e.g., Wudao, S2ORC), enabling the model to handle varying lengths and information densities. Second, it improves \textit{domain generalization} by combining formal sources (papers, news) with informal sources (forums, Q\&A), reducing style overfitting. Third, the large multilingual portion acts as an \textit{alignment anchor}, allowing English- and Chinese-specific knowledge to transfer more effectively to lower-resource languages.
\begin{table}[htbp]
  \centering
  \caption{List of pre-training data sources. The list now includes SkyPile and Falcon datasets.}
  \label{tab:pretrain_data_list}
  \renewcommand{\arraystretch}{1.2} 
  \setlength{\tabcolsep}{12pt}      
  \footnotesize
  \begin{tabular}{llr} 
    \toprule
    \textbf{Source} & \textbf{Language} & \textbf{Pairs} \\
    \midrule
    \href{https://huggingface.co/datasets/McAuley-Lab/Amazon-Reviews-2023}{Amazon-Reviews}~\citep{DBLP:journals/corr/abs-2403-03952} & Multilingual & 23M \\
    \href{https://huggingface.co/datasets/intfloat/multilingual_cc_news}{CC-News}~\citep{DBLP:conf/isiwi/HamborgMBG17} & Multilingual & 100M \\
    \href{https://huggingface.co/datasets/allenai/nllb}{NLLB}~\citep{DBLP:journals/corr/abs-2207-04672} & Multilingual & 2M \\
    \href{https://huggingface.co/datasets/Cohere/wikipedia-2023-11-embed-multilingual-v3}{Wikipedia}~\citep{wikidump} & Multilingual & 100M \\
    \href{https://huggingface.co/datasets/bigscience/xP3}{xP3}~\citep{DBLP:conf/acl/MuennighoffWSRB23} & Multilingual & 19M \\
    \href{https://huggingface.co/datasets/GEM/xlsum}{XL-Sum}~\citep{DBLP:conf/acl/HasanBIMLKRS21} & Multilingual & 1M \\
    \href{https://huggingface.co/datasets/nthakur/swim-ir-monolingual}{SWIM-IR (Mono)}~\citep{DBLP:conf/naacl/ThakurNAWLC24} & Multilingual & 3M \\
    \href{https://huggingface.co/datasets/nthakur/swim-ir-cross-lingual}{SWIM-IR (Cross)}~\citep{DBLP:conf/naacl/ThakurNAWLC24} & Multilingual & 15M \\
    \midrule
    \href{https://huggingface.co/datasets/neuclir/csl}{CSL}~\citep{DBLP:conf/coling/LiZ0S0MZ22} & Chinese & 0.4M \\
    \href{https://data.baai.ac.cn/details/WuDaoCorporaText}{Wudao}~\citep{DBLP:journals/aiopen/YuanZDDLCZYT21} & Chinese & 44M \\
    \href{https://huggingface.co/datasets/Skywork/SkyPile-150B}{SkyPile} & Chinese & 81M \\ 
    \href{https://huggingface.co/datasets/SirlyDreamer/THUCNews}{THUCNews}~\citep{thuctc} & Chinese & 0.8M \\
    \href{https://huggingface.co/datasets/wangrui6/Zhihu-KOL}{Zhihu-KOL} & Chinese & 0.8M \\
    \midrule
    \href{https://huggingface.co/datasets/sentence-transformers/codesearchnet}{CodeSearchNet}~\citep{DBLP:journals/corr/abs-1909-09436} & English & 1M \\
    \href{https://huggingface.co/datasets/tiiuae/falcon-refinedweb}{Falcon} & English & 128M \\ 
    \href{https://huggingface.co/datasets/sentence-transformers/paq}{PAQ}~\citep{DBLP:journals/tacl/LewisWLMKPSR21} & English & 9M \\
    \href{https://huggingface.co/datasets/sentence-transformers/reddit}{Reddit} & English & 100M \\
    \href{https://huggingface.co/datasets/teven/stackexchange}{StackExchange} & English & 14M \\
    \href{https://huggingface.co/datasets/sentence-transformers/s2orc}{S2ORC} & English & 41M \\
    \bottomrule
  \end{tabular}
\end{table}

\subsection{Fine-tuning Data Curation}
\label{sec:finetuning_data}

To align the pre-trained model with instruction-following behavior and downstream objectives, we construct a fine-tuning mixture spanning over 70 high-quality corpora. Table~\ref{tab:Fine-tuning_data_list} summarizes the datasets, which cover three primary task families: retrieval, semantic textual similarity (STS), and classification. The curation balances English, Chinese, and multilingual coverage while preserving task heterogeneity.

Retrieval data constitutes the core of the fine-tuning stage, focusing on query--passage matching (s2p). We include established English benchmarks such as \textit{MSMARCO}~\citep{DBLP:conf/nips/NguyenRSGTMD16} and \textit{Natural Questions}, together with large-scale Chinese retrieval corpora such as \textit{T2Ranking}~\citep{DBLP:conf/sigir/XieDWLYG0LL0M23} and \textit{mMARCO}. For cross-lingual retrieval, we incorporate \textit{MIRACL} and \textit{Mr. TyDi}, which provide high-quality query--passage pairs across diverse languages.

To improve sensitivity to fine-grained semantic nuances, we add STS-style datasets including \textit{STS Benchmark}, \textit{Quora Question Pairs}, and \textit{AFQMC}. These typically follow an s2s format and encourage the model to separate semantically equivalent from non-equivalent pairs.

We further include classification and clustering corpora (e.g., \textit{AmazonPolarity}, \textit{TNews}, \textit{Reddit-Clustering}) to encourage linearly separable semantic structure in the embedding space, improving representation geometry beyond pure pair matching.

Quality control is critical at this stage. As indicated by the ``Pairs (filtered)'' column in Table~\ref{tab:Fine-tuning_data_list}, we apply a filtering pipeline that removes duplicates, filters low-overlap positives, and discards samples that are too short or noisy. For example, in \textit{Reddit-P2P}, aggressive filtering reduces the data from $\sim$12M to $\sim$42k pairs, mitigating noise-induced degeneration and improving training efficiency.
\begin{longtable}{llcrr}
    \caption{List of fine-tuning datasets. The datasets are grouped by language and categorized by task type (Retrieval, STS, Classification). Columns show the original pair count and the count after filtering.} 
    \label{tab:Fine-tuning_data_list} \\
    
    \toprule
    \textbf{Source} & \textbf{Type} & \textbf{Categ.} & \textbf{Pairs} & \textbf{Pairs (Filt.)} \\
    \midrule
    \endfirsthead
    
    \caption[]{Fine-tuning data list (continued)} \\
    \toprule
    \textbf{Source} & \textbf{Type} & \textbf{Categ.} & \textbf{Pairs} & \textbf{Pairs (Filt.)} \\
    \midrule
    \endhead
    
    \bottomrule
    \multicolumn{5}{r}{\textit{Continued on next page...}} \\
    \endfoot
    \bottomrule
    \endlastfoot

    \multicolumn{5}{l}{\cellcolor{gray!10}\textbf{\textit{English Resources}}} \\
    \addlinespace[0.5ex]

    \href{https://huggingface.co/datasets/m-a-p/CodeFeedback-Filtered-Instruction}{CodeFeedback}~\citep{DBLP:conf/acl/ZhengZSLLFCY24} & Retrieval & s2p & 50,000 & 49,090 \\
    \href{https://huggingface.co/datasets/rusano/ELI5_custom}{ELI5}~\citep{DBLP:conf/acl/FanJPGWA19} & Retrieval & s2p & 100,000 & 76,408 \\
    \href{https://github.com/chaitanyamalaviya/ExpertQA}{ExpertQA}~\citep{DBLP:conf/naacl/MalaviyaLCSYR24} & Retrieval & s2p & 1,261 & 1,252 \\
    \href{https://github.com/allenai/gooaq}{GooAQ}~\citep{DBLP:conf/emnlp/KhashabiNKSHC21} & Retrieval & s2p & 50,000 & 49,833 \\
    \href{https://hf.co/datasets/GritLM/MEDI2BGE}{MEDI2BGE}~\citep{DBLP:journals/corr/abs-2402-09906} & Retrieval & s2p & 100,000 & 71,790 \\
    \href{https://huggingface.co/datasets/Open-Orca/OpenOrca}{OpenOrca}~\citep{DBLP:journals/corr/abs-2306-02707} & Retrieval & s2p & 40,000 & 38,623 \\
    \href{https://huggingface.co/datasets/sentence-transformers/paq}{PAQ}~\citep{DBLP:journals/tacl/LewisWLMKPSR21} & Retrieval & s2p & 50,000 & 49,849 \\
    \href{https://huggingface.co/datasets/qiaojin/PubMedQA}{PubMedQA}~\citep{DBLP:conf/emnlp/JinDLCL19} & Retrieval & s2p & 80,000 & 79,954 \\
    \href{https://huggingface.co/datasets/kyunghyuncho/search_qa}{SearchQA}~\citep{DBLP:journals/corr/DunnSHGCC17} & Retrieval & s2p & 10,000 & 9,988 \\
    \href{https://huggingface.co/datasets/TitanMLData/arxiv_qa}{arxiv\_qa} & Retrieval & s2p & 23,397 & 17,927 \\
    \href{https://huggingface.co/datasets/intfloat/multilingual_cc_news}{CC-News}~\citep{DBLP:conf/isiwi/HamborgMBG17} & Retrieval & s2p & 30,000 & 28,246 \\
    \href{https://huggingface.co/datasets/irds/cord19_trec-covid}{TREC-COVID}~\citep{DBLP:journals/sigir/VoorheesABDHLRS20} & Retrieval & s2p & 50,000 & 48,517 \\
    \href{https://huggingface.co/datasets/BeIR/dbpedia-entity-generated-queries}{DBpedia-Entity}~\citep{DBLP:conf/nips/Thakur0RSG21} & Retrieval & s2p & 100,000 & 96,792 \\
    \href{https://huggingface.co/datasets/tasksource/esci}{ESCI}~\citep{DBLP:journals/corr/abs-2206-06588} & Retrieval & s2p & 30,000 & 26,043 \\
    \href{https://huggingface.co/datasets/maxzoech/fever}{FEVER}~\citep{DBLP:conf/naacl/ThorneVCM18} & Retrieval & s2p & 87,855 & 87,216 \\
    \href{https://huggingface.co/datasets/irds/beir_fiqa_train}{FiQA}~\citep{DBLP:conf/www/MaiaHFDMZB18} & Retrieval & s2p & 5,490 & 4,689 \\
    \href{https://huggingface.co/datasets/hotpotqa/hotpot_qa}{HotpotQA}~\citep{DBLP:conf/emnlp/Yang0ZBCSM18} & Retrieval & s2p & 184,057 & 150,153 \\
    \href{https://huggingface.co/datasets/Shitao/MLDR}{MLDR}~\citep{DBLP:journals/corr/abs-2402-03216} & Retrieval & s2p & 41,434 & 31,097 \\
    \href{https://huggingface.co/datasets/Tevatron/msmarco-passage}{MSMARCO}~\citep{DBLP:conf/nips/NguyenRSGTMD16} & Retrieval & s2p & 175,133 & 174,190 \\
    \href{https://huggingface.co/datasets/mteb/msmarco-v2}{MSMARCO-v2}~\citep{DBLP:conf/nips/NguyenRSGTMD16} & Retrieval & s2p & 277,144 & 258,617 \\
    \href{https://huggingface.co/datasets/BeIR/nfcorpus-generated-queries}{NFCorpus}~\citep{DBLP:conf/ecir/BotevaGSR16} & Retrieval & s2p & 10,824 & 10,471 \\
    \href{https://huggingface.co/datasets/neural-bridge/rag-dataset-12000}{rag-dataset-12k} & Retrieval & s2p & 9,590 & 9,272 \\
    \href{https://huggingface.co/datasets/Tevatron/scifact}{SciFact}~\citep{DBLP:conf/emnlp/WaddenLLWZCH20} & Retrieval & s2p & 809 & 794 \\
    \href{https://huggingface.co/datasets/rajpurkar/squad_v2}{SQuAD 2.0}~\citep{DBLP:conf/acl/RajpurkarJL18} & Retrieval & s2p & 130,217 & 125,816 \\
    \href{https://huggingface.co/datasets/multi-train/emb-triviaqa-train}{TriviaQA}~\citep{DBLP:conf/acl/JoshiCWZ17} & Retrieval & s2p & 52,886 & 44,442 \\
    \href{https://huggingface.co/datasets/openai/webgpt_comparisons}{WebGPT Comp.}~\citep{DBLP:journals/corr/abs-2112-09332} & Retrieval & s2p & 19,242 & 18,924 \\
    \href{https://huggingface.co/datasets/Tevatron/wikipedia-nq}{Natural Questions}~\citep{DBLP:journals/tacl/KwiatkowskiPRCP19} & Retrieval & s2p & 58,622 & 56,377 \\
    \href{https://huggingface.co/datasets/sentence-transformers/yahoo-answers}{Yahoo Answers} & Retrieval & s2p & 30,000 & 21,724 \\
    \href{http://nlp.cis.unimelb.edu.au/resources/cqadupstack/}{CQADupStack}~\citep{DBLP:conf/adcs/HoogeveenVB15} & Retrieval & s2p & 24,045 & 7,356 \\
    \href{https://huggingface.co/datasets/kiddothe2b/contract-nli}{ContractNLI}~\citep{DBLP:conf/emnlp/KoreedaM21} & STS & s2s & 3,195 & 628 \\
    \href{https://huggingface.co/datasets/SetFit/mnli}{MultiNLI}~\citep{DBLP:conf/naacl/WilliamsNB18} & STS & s2s & 64,674 & 63,701 \\
    \href{https://huggingface.co/datasets/breakend/nllb-multi-domain}{NLLB}~\citep{DBLP:journals/corr/abs-2207-04672} & STS & s2s & 36,000 & 26,504 \\
    \href{https://huggingface.co/datasets/sentence-transformers/embedding-training-data}{Quora}~\citep{quora-question-pairs} & STS & s2s & 92,674 & 89,558 \\
    \href{https://huggingface.co/datasets/multi-train/WikiAnswers_1107}{WikiAnswers}~\citep{DBLP:conf/kdd/FaderZE14} & STS & s2s & 50,000 & 47,686 \\
    \href{https://huggingface.co/datasets/JeremiahZ/simcse_sup_nli}{SimCSE NLI}~\citep{DBLP:conf/emnlp/GaoYC21} & STS & s2s & 252,397 & 217,099 \\
    \href{https://huggingface.co/datasets/stanfordnlp/snli}{SNLI}~\citep{DBLP:conf/emnlp/BowmanAPM15} & STS & s2s & 24,686 & 16,480 \\
    \href{https://huggingface.co/datasets/mteb/raw_arxiv}{arXiv} & Classification & s2s, p2s & 15,000 & 14,529 \\
    \href{https://huggingface.co/datasets/mteb/raw_biorxiv}{Biorxiv} & Classification & s2s, p2s & 6,862 & 6,787 \\
    \href{https://huggingface.co/datasets/mteb/raw_medrxiv}{Medrxiv} & Classification & s2s, p2s & 2,012 & 1,999 \\
    \href{https://github.com/UKPLab/TWEAC-qa-agent-selection/tree/master/data/reddit/train}{Reddit-Clustering} & Classification & s2s & 128,000 & 25,600 \\
    \href{https://huggingface.co/datasets/sentence-transformers/reddit-title-body}{Reddit-P2P} & Classification & p2s & 12,704,958 & 42,480 \\
    \href{https://github.com/UKPLab/TWEAC-qa-agent-selection/tree/master/data/stackexchange/train}{Stack-Clustering} & Classification & s2s & 1,014,826 & 50,530 \\
    \href{https://huggingface.co/datasets/flax-sentence-embeddings/stackexchange_title_body_jsonl}{Stack-P2P} & Classification & p2s & 25,333,327 & 48,800 \\
    \href{https://scikit-learn.org/0.19/datasets/twenty_newsgroups.html}{20Newsgroups} & Classification & s2s & 11,314 & 6,233 \\
    \href{https://huggingface.co/datasets/mteb/amazon_polarity}{AmazonPolarity} & Classification & s2s & 10,000 & 9,007 \\
    \href{https://huggingface.co/datasets/mteb/imdb}{IMDB} & Classification & s2s & 10,000 & 8,575 \\
    \href{https://huggingface.co/datasets/mteb/banking77}{Banking77} & Classification & s2s & 10,000 & 9,937 \\
    \href{https://huggingface.co/datasets/mteb/emotion}{Emotion} & Classification & s2s & 10,000 & 10,000 \\
    \href{https://huggingface.co/datasets/mteb/tweet_sentiment_extraction}{TweetSentiment} & Classification & s2s & 10,000 & 10,000 \\
    \href{https://huggingface.co/datasets/mteb/toxic_conversations_50k}{ToxicConversations} & Classification & s2s & 7,916 & 7,800 \\

    \addlinespace
    \multicolumn{5}{l}{\cellcolor{gray!10}\textbf{\textit{Chinese Resources}}} \\
    \addlinespace[0.5ex]

    \href{https://huggingface.co/datasets/shibing624/AdvertiseGen}{AdvertiseGen} & Retrieval & s2p & 20,000 & 17,526 \\
    \href{https://www.luge.ai/\#/luge/dataDetail?id=44}{CHEF} & Retrieval & s2p & 4,952 & 4,824 \\
    \href{https://huggingface.co/datasets/michaelwzhu/ChatMed_Consult_Dataset}{ChatMed} & Retrieval & s2p & 20,000 & 18,608 \\
    \href{https://huggingface.co/datasets/erhwenkuo/squad-cmrc2018-zhtw}{CMRC 2018} & Retrieval & s2p & 10,000 & 9,753 \\
    \href{https://huggingface.co/datasets/voidful/DRCD}{DRCD} & Retrieval & s2p & 5,000 & 4,714 \\
    \href{https://huggingface.co/datasets/hugcyp/LCSTS}{LCSTS} & Retrieval & s2p & 20,000 & 19,535 \\
    \href{https://huggingface.co/datasets/paralym/lima-chinese}{LIMA} & Retrieval & s2p & 2,058 & 1,991 \\
    \href{https://github.com/Alibaba-NLP/Multi-CPR}{Multi-CPR} & Retrieval & s2p & 287,881 & 234,587 \\
    \href{https://huggingface.co/datasets/C-MTEB/PAWSX}{PAWS-X (zh)} & Retrieval & s2p & 49,401 & 19,289 \\
    \href{https://github.com/sufengniu/RefGPT/blob/main/README_EN.md}{RefGPT} & Retrieval & s2p & 50,000 & 49,896 \\
    \href{https://huggingface.co/datasets/THUIR/T2Ranking}{T2Ranking} & Retrieval & s2p & 199,412 & 188,606 \\
    \href{https://huggingface.co/datasets/SirlyDreamer/THUCNews}{THUCNews} & Retrieval & s2p & 20,000 & 19,288 \\
    \href{https://www.luge.ai/\#/luge/dataDetail?id=62}{UMETRIP-QA} & Retrieval & s2p & 2,647 & 2,537 \\
    \href{https://github.com/thunlp/WebCPM}{WebCPM} & Retrieval & s2p & 1,605 & 1,602 \\
    \href{https://www.datafountain.cn/competitions/424/datasets}{cCOVID-News} & Retrieval & s2p & 5,000 & 4,727 \\
    \href{https://huggingface.co/datasets/wangrongsheng/cMedQA-V2.0}{cMedQA-V2.0} & Retrieval & s2p & 223,851 & 88,109 \\
    \href{https://huggingface.co/datasets/neuclir/csl}{CSL} & Retrieval & s2p & 20,000 & 19,945 \\
    \href{https://huggingface.co/datasets/sentence-transformers/dureader}{DuReader} & Retrieval & s2p & 80,416 & 79,229 \\
    \href{https://huggingface.co/datasets/luozhouyang/dureader}{DuReader\textsubscript{checklist}} & Retrieval & s2p & 99,992 & 97,764 \\
    \href{https://huggingface.co/datasets/sentence-transformers/law-gpt}{law-gpt} & Retrieval & s2p & 500 & 500 \\
    \href{https://www.heywhale.com/mw/dataset/5e953ca8e7ec38002d02fca7/content}{lawzhidao} & Retrieval & s2p & 8,000 & 6,784 \\
    \href{https://huggingface.co/datasets/unicamp-dl/mmarco}{mMARCO (zh)} & Retrieval & s2p & 400,000 & 379,870 \\
    \href{https://huggingface.co/datasets/infgrad/retrieval_data_llm}{retrieval\_data\_llm} & Retrieval & s2p & 32,768 & 32,551 \\
    \href{https://huggingface.co/datasets/suolyer/webqa}{webqa} & Retrieval & s2p & 5,000 & 4,988 \\
    \href{https://huggingface.co/datasets/C-MTEB/AFQMC}{AFQMC} & STS & s2s & 4,041 & 3,876 \\
    \href{https://huggingface.co/datasets/C-MTEB/ATEC}{ATEC} & STS & s2s & 62,477 & 11,387 \\
    \href{https://huggingface.co/datasets/C-MTEB/BQ}{BQ} & STS & s2s & 100,000 & 10,000 \\
    \href{https://github.com/china-ai-law-challenge/CAIL2019/tree/master/scm}{CAIL2019-SCM} & STS & s2s & 5,102 & 648 \\
    \href{https://www.luge.ai/\#/luge/dataDetail?id=39}{CINLID} & STS & s2s & 5,000 & 2,883 \\
    \href{https://github.com/IAdmireu/ChineseSTS}{ChineseSTS} & STS & s2s & 2,500 & 2,497 \\
    \href{https://huggingface.co/datasets/fenffef/cmnli}{CMNLI} & STS & s2s & 125,356 & 119,029 \\
    \href{https://huggingface.co/datasets/shibing624/nli_zh}{nli\_zh} & STS & s2s & 218,887 & 185,787 \\
    \href{https://huggingface.co/datasets/Fred666/ocnli}{OCNLI} & STS & s2s & 13,464 & 11,937 \\
    \href{https://github.com/CLUEbenchmark/QBQTC/tree/main}{QBQTC} & STS & s2s & 51,620 & 47,223 \\
    \href{https://github.com/CLUEbenchmark/SimCLUE}{SimCLUE} & STS & s2s & 344,038 & 290,699 \\
    \href{https://huggingface.co/datasets/xnli}{XNLI (zh)} & STS & s2s & 80,000 & 74,252 \\
    \href{https://huggingface.co/datasets/neuclir/csl}{CSL} & Classification & s2s, p2s & 15,000 & 12,249 \\
    \href{https://huggingface.co/datasets/SirlyDreamer/THUCNews}{THUCNews} & Classification & s2s & 10,000 & 9,690 \\
    \href{https://huggingface.co/datasets/fenffef/tnews}{TNews} & Classification & s2s & 10,000 & 6,762 \\
    \href{https://huggingface.co/datasets/C-MTEB/JDReview-classification}{JDReview} & Classification & s2s & 1,232 & 1,232 \\
    \href{https://huggingface.co/datasets/fenffef/iflytek}{IFlyTek} & Classification & s2s & 10,000 & 8,221 \\
    \href{https://huggingface.co/datasets/C-MTEB/OnlineShopping-classification}{OnlineShopping} & Classification & s2s & 7,852 & 7,600 \\
    \href{https://huggingface.co/datasets/C-MTEB/waimai-classification}{Waimai} & Classification & s2s & 7,384 & 7,376 \\

    \addlinespace
    \multicolumn{5}{l}{\cellcolor{gray!10}\textbf{\textit{Multilingual Resources}}} \\
    \addlinespace[0.5ex]

    \href{https://huggingface.co/datasets/CohereForAI/aya_dataset}{Aya Dataset}~\citep{DBLP:conf/acl/SinghVD0MKSPMOZ24} & Retrieval & s2p & 30,000 & 26,292 \\
    \href{https://huggingface.co/datasets/sentence-transformers/miracl}{MIRACL}~\citep{DBLP:journals/tacl/0018TOKAL0RL23} & Retrieval & s2p & 40,151 & 39,946 \\
    \href{https://huggingface.co/datasets/castorini/mr-tydi}{Mr. TyDi}~\citep{DBLP:journals/corr/abs-2108-08787} & Retrieval & s2p & 48,729 & 46,997 \\
    \href{https://huggingface.co/datasets/maximedb/paws-x-all}{PAWS-X}~\citep{DBLP:conf/emnlp/YangZTB19} & STS & s2s & 128,435 & 128,398 \\
    \href{https://huggingface.co/datasets/mteb/amazon_reviews_multi}{AmazonReviews} & Classification & s2s & 10,000 & 7,721 \\
    \href{https://huggingface.co/datasets/mteb/amazon_counterfactual}{AmazonCounterfactual} & Classification & s2s & 10,000 & 8,323 \\
    \href{https://huggingface.co/datasets/mteb/multilingual-sentiment-classification}{MultiSentiment} & Classification & s2s & 10,000 & 9,804 \\
    \href{https://huggingface.co/datasets/mteb/amazon_massive_intent}{Amazon Massive} & Classification & s2s & 10,000 & 7,832 \\
    \href{https://huggingface.co/datasets/mteb/amazon_massive_scenario}{Amazon Scenario} & Classification & s2s & 10,000 & 7,078 \\
    \href{https://huggingface.co/datasets/mteb/mtop_domain}{MTOPDomain} & Classification & s2s & 10,000 & 9,610 \\
    \href{https://huggingface.co/datasets/mteb/mtop_intent}{MTOPIntent} & Classification & s2s & 10,000 & 7,952 \\

\end{longtable}

\subsection{Unified Instruction Templates for MTEB Evaluation}

Table~\ref{tab:task_instruction_detailed_list} summarizes the instruction templates used across MTEB (eng, v2). Although MTEB covers heterogeneous capabilities---classification, clustering, semantic similarity, retrieval, reranking, and summarization---a large fraction of tasks can be expressed using a small family of instruction patterns. This enables a unified instruction-driven evaluation without per-dataset prompt engineering.

For label-centric tasks such as \textit{Classification} and \textit{Clustering}, we use task-specific natural language descriptions that encode dataset semantics (e.g., emotion, intent, topic). This formulation maps the original objective into a consistent ``Instruct/Query'' interface.

For semantic matching tasks---including \textit{Retrieval}, \textit{Reranking}, \textit{Pair Classification}, and \textit{STS}---we use shared templates due to their closely aligned objectives. 
As indicated by the asterisks in Table~\ref{tab:task_instruction_detailed_list}, the same instruction is applied to all remaining tasks of each type. 
Only a small number of datasets require overrides to better reflect dataset intent (e.g., duplicate question detection in QuoraRetrieval or StackExchange-style tasks).

This unified design offers two practical benefits: it reduces the mismatch between instruction-tuned behavior and embedding-style evaluation, and it provides a principled way to inject task intent into continuous representations, yielding a more standardized and interpretable evaluation pipeline.
\begin{table}[htbp]
  \centering
  \caption{Detailed task instruction list for MTEB evaluation. Pair Classification$^*$, Reranking$^*$, Retrieval$^*$, and STS$^*$ indicate we use the same instructions for all the respective remaining tasks.}
  \label{tab:task_instruction_detailed_list}
  \renewcommand{\arraystretch}{1.3} 
  \tiny
  \begin{tabularx}{\textwidth}{lX} 
    \toprule
    \textbf{Task Name} & \textbf{Instruction} \\
    \midrule
    
    \multicolumn{2}{c}{\textit{\textbf{Classification}}} \\
    \addlinespace[0.5ex] 
    
    AmazonCounterfactual & Instruct: Given an Amazon review, judge whether it is counterfactual. \textbackslash n Query: \{query\} \\
    AmazonPolarity & Instruct: Classify Amazon reviews into positive or negative sentiment. \textbackslash n Query: \{query\} \\
    AmazonReviews & Instruct: Classify the given Amazon review into its appropriate rating category. \textbackslash n Query: \{query\} \\
    Banking77 & Instruct: Given an online banking query, find the corresponding intents. \textbackslash n Query: \{query\} \\
    Emotion & Instruct: Classify the emotion expressed in the given Twitter message into one of the six emotions: anger, fear, joy, love, sadness, and surprise. \textbackslash n Query: \{query\} \\
    Imdb & Instruct: Classify the sentiment expressed in the given movie review text from the IMDB dataset. \textbackslash n Query: \{query\} \\
    MassiveIntent & Instruct: Given a user utterance as query, find the user intents. \textbackslash n Query: \{query\} \\
    MassiveScenario & Instruct: Given a user utterance as query, find the user scenarios. \textbackslash n Query: \{query\} \\
    MTOPDomain & Instruct: Classify the intent domain of the given utterance in task-oriented conversation. \textbackslash n Query: \{query\} \\
    MTOPIntent & Instruct: Classify the intent of the given utterance in task-oriented conversation. \textbackslash n Query: \{query\} \\
    ToxicConversations & Instruct: Classify the given comments as either toxic or not toxic. \textbackslash n Query: \{query\} \\
    TweetSentiment & Instruct: Classify the sentiment of a given tweet as either positive, negative, or neutral. \textbackslash n Query: \{query\} \\
    TNews & Instruct: Categorize the given news title. \textbackslash n Query: \{query\} \\
    IFlyTek & Instruct: Given an App description text, find the appropriate fine-grained category. \textbackslash n Query: \{query\} \\
    MultilingualSentiment & Instruct: Classify sentiment of the customer review into positive, neutral, or negative. \textbackslash n Query: \{query\} \\
    JDReview & Instruct: Classify sentiment of the customer review for iPhone into positive or negative. \textbackslash n Query: \{query\} \\
    OnlineShopping & Instruct: Classify sentiment of the customer review into positive or negative. \textbackslash n Query: \{query\} \\
    Waimai & Instruct: Classify the customer review from a food takeaway platform into positive or negative. \textbackslash n Query: \{query\} \\

    \midrule
    \multicolumn{2}{c}{\textit{\textbf{Clustering}}} \\
    \addlinespace[0.5ex]

    ArxivClusteringP2P & Instruct: Identify the main and secondary category of Arxiv papers based on the titles and abstracts. \textbackslash n Query: \{query\} \\
    ArxivClusteringS2S & Instruct: Identify the main and secondary category of Arxiv papers based on the titles. \textbackslash n Query: \{query\} \\
    BiorxivClusteringP2P & Instruct: Identify the main category of Biorxiv papers based on the titles and abstracts. \textbackslash n Query: \{query\} \\
    BiorxivClusteringS2S & Instruct: Identify the main category of Biorxiv papers based on the titles. \textbackslash n Query: \{query\} \\
    MedrxivClusteringP2P & Instruct: Identify the main category of Medrxiv papers based on the titles and abstracts. \textbackslash n Query: \{query\} \\
    MedrxivClusteringS2S & Instruct: Identify the main category of Medrxiv papers based on the titles. \textbackslash n Query: \{query\} \\
    RedditClustering & Instruct: Identify the topic or theme of Reddit posts based on the titles. \textbackslash n Query: \{query\} \\
    RedditClusteringP2P & Instruct: Identify the topic or theme of Reddit posts based on the titles and posts. \textbackslash n Query: \{query\} \\
    StackExchangeClustering & Instruct: Identify the topic or theme of StackExchange posts based on the titles. \textbackslash n Query: \{query\} \\
    StackExchangeP2P & Instruct: Identify the topic or theme of StackExchange posts based on the given paragraphs. \textbackslash n Query: \{query\} \\
    TwentyNewsgroups & Instruct: Identify the topic or theme of the given news articles. \textbackslash n Query: \{query\} \\
    CLSClusteringS2S & Instruct: Identify the main category of scholar papers based on the titles. \textbackslash n Query: \{query\} \\
    CLSClusteringP2P & Instruct: Identify the main category of scholar papers based on the titles and abstracts. \textbackslash n Query: \{query\} \\
    ThuNewsClusteringS2S & Instruct: Identify the topic or theme of the given news articles based on the titles. \textbackslash n Query: \{query\} \\
    ThuNewsClusteringP2P & Instruct: Identify the topic or theme of the given news articles based on the titles and contents. \textbackslash n Query: \{query\} \\

    \midrule
    \multicolumn{2}{c}{\textit{\textbf{Pair Classification}}} \\
    \addlinespace[0.5ex]
    
    Pair Classification$^*$ & Instruct: Retrieve semantically similar text. \textbackslash n Query: \{query\} \\
    SprintDuplicateQuestions & Instruct: Retrieve semantically similar questions. \textbackslash n Query: \{query\} \\

    \midrule
    \multicolumn{2}{c}{\textit{\textbf{Reranking}}} \\
    \addlinespace[0.5ex]
    
    Reranking$^*$ & Instruct: Given a query, retrieve documents that answer the query. \textbackslash n Query: \{query\} \\
    AskUbuntuDupQuestions & Instruct: Retrieve semantically similar questions. \textbackslash n Query: \{query\} \\
    StackOverflowDupQuestions & Instruct: Retrieve semantically similar questions. \textbackslash n Query: \{query\} \\
    SciDocsRR & Instruct: Retrieve relevant paper titles. \textbackslash n Query: \{query\} \\

    \midrule
    \multicolumn{2}{c}{\textit{\textbf{Retrieval}}} \\
    \addlinespace[0.5ex]
    
    Retrieval$^*$ & Instruct: Given a query, retrieve documents that answer the query. \textbackslash n Query: \{query\} \\
    QuoraRetrieval & Instruct: Retrieve semantically similar questions. \textbackslash n Query: \{query\} \\
    CQADupstack & Instruct: Given a question, retrieve detailed question descriptions from StackExchange that are duplicates to the given question. \textbackslash n Query: \{query\} \\

    \midrule
    \multicolumn{2}{c}{\textit{\textbf{STS}}} \\
    \addlinespace[0.5ex]
    
    STS$^*$ & Instruct: Retrieve semantically similar text. \textbackslash n Query: \{query\} \\

    \midrule
    \multicolumn{2}{c}{\textit{\textbf{Summarization}}} \\
    \addlinespace[0.5ex]
    
    SummEval & Instruct: Retrieve semantically similar summaries. \textbackslash n Query: \{query\} \\
    
    \bottomrule
  \end{tabularx}
\end{table}

\end{document}